\documentclass[10pt,twocolumn,letterpaper]{article}

\usepackage{multirow}
\usepackage{wacv}
\usepackage{times}
\usepackage{epsfig}
\usepackage{graphicx}
\usepackage{amsmath}
\usepackage{amssymb}
\usepackage{soul}
\usepackage{array}
\usepackage{ltablex}
\usepackage{supertabular,booktabs}
\usepackage{float}
\usepackage{subcaption}
\restylefloat{table}
\usepackage{rotating}
\usepackage{multicol}
\usepackage{bold-extra}
\usepackage{fancyhdr}
\usepackage{hyperref}
\wacvfinalcopy 

\def\wacvPaperID{****} 

\ifwacvfinal\fi
\begin{document}
\title{\bfseries{\scshape{SAVOIAS}}: A Diverse, Multi-Category Visual Complexity Dataset}
 \vspace{-0.95cm}

\author{
Elham Saraee,
Mona Jalal,
Margrit Betke \\
 \vspace{-0.15cm}
{\large Department of Computer Science, Boston University, USA}\\
\\
{ \small esaraee@bu.edu, jalal@bu.edu, betke@bu.edu}\\
{\small \href{https://www.bu.edu/cs/ivc/}{https://www.bu.edu/cs/ivc/}}}
\maketitle
\fancypagestyle{plain}{
\fancyhf{} 
\fancyfoot[L]{\footnotesize{
The copyright of this paper and the annotations remain with the authors.  The copyright of the images shown belongs to the image owners. The Savoias dataset is publicly available at  {\it https://github.com/esaraee/Savoias-Dataset.}
}}
\fancyfoot[C]{}
\fancyfoot[R]{}

\renewcommand{\headrulewidth}{0pt}
\renewcommand{\footrulewidth}{0pt}
}

 \vspace{-1.75cm}

\begin{abstract}
 \vspace{-0.35cm}

Visual complexity identifies the level of intricacy and details in an image or the level of difficulty to describe the image. It is an important concept in a variety of areas such as cognitive psychology, computer vision and visualization, and advertisement. Yet, efforts to create large, downloadable image datasets with diverse content and unbiased groundtruthing are lacking.  In this work, we introduce {\sc Savoias} a visual complexity dataset that compromises of more than 1,400 images from seven image categories relevant to the above research areas, namely \textit{\textbf{S}cenes}, \textit{\textbf{A}dvertisements}, \textit{\textbf{V}isualization} and infographics, \textit{\textbf{O}bjects}, \textit{\textbf{I}nterior design}, \textit{\textbf{A}rt}, and \textit{\textbf{S}uprematism}. 
The images in each category portray diverse characteristics including various low-level and high-level features, objects, backgrounds, textures and patterns, text, and graphics. 
The ground truth for {\sc Savoias} is obtained by crowdsourcing more than 37,000 pairwise comparisons of images using the forced-choice methodology and with more than 1,600 contributors. The resulting relative scores are then converted to absolute visual complexity scores using the Bradley-Terry method and matrix completion. When applying five state-of-the-art algorithms to analyze the visual complexity of the images in the {\sc Savoias} dataset, we found that the scores obtained from these baseline tools only correlate well with crowdsourced labels for abstract patterns in the Suprematism category (Pearson correlation r=0.84).   For the other categories, in particular, the objects and advertisement categories, low correlation coefficients were revealed (r=0.3 and 0.56, respectively).  These findings suggest that (1) state-of-the-art approaches are mostly insufficient and (2) {\sc Savoias} enables category-specific method development, which is likely to improve the impact of visual complexity analysis on specific application areas, including computer vision.

\end{abstract}

\begin{figure}[h]
  \centering
  \hspace{-0.25em}
    {\includegraphics[height=.48\linewidth,width=.48\linewidth]{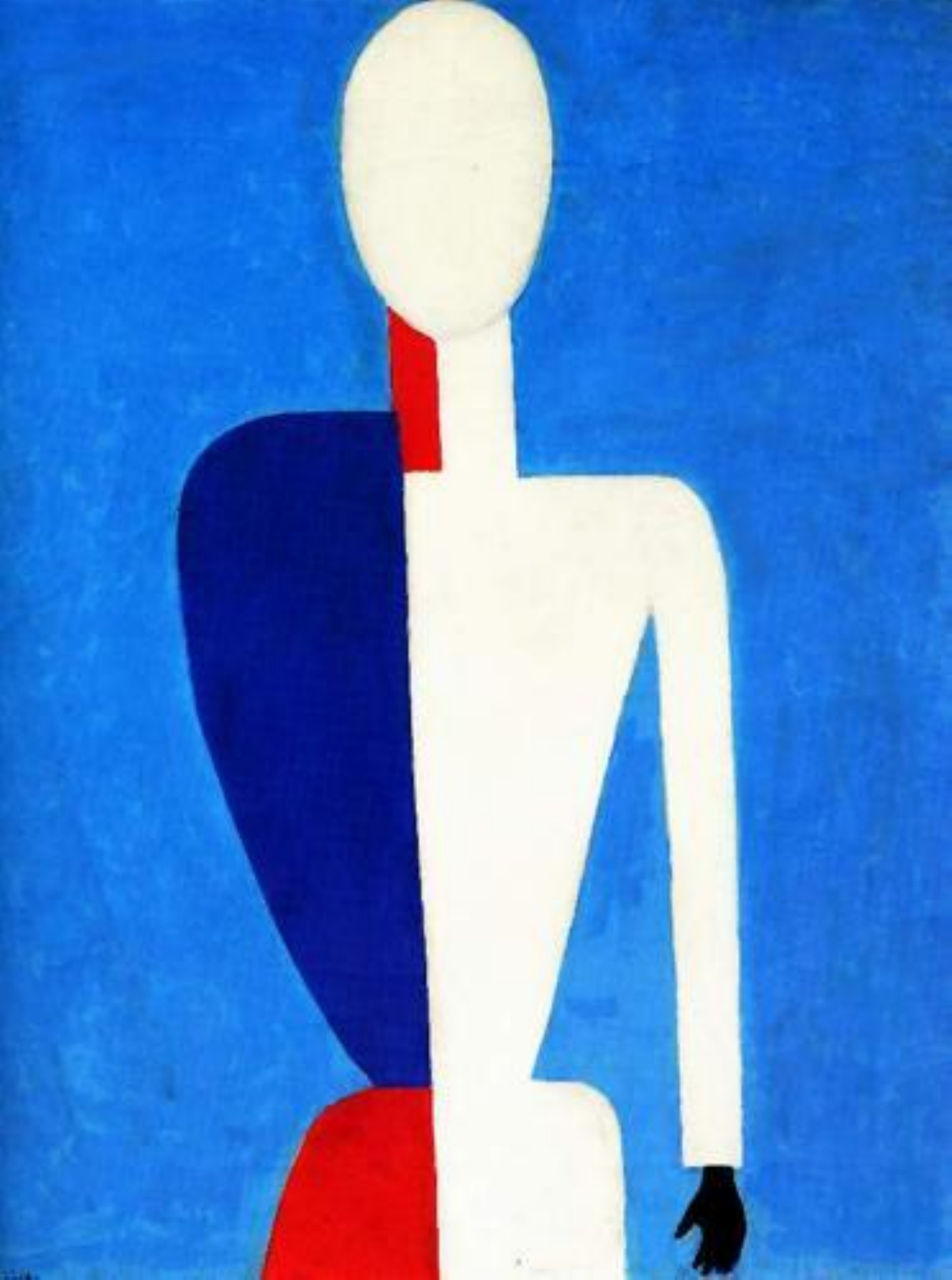}}
    {\includegraphics[height=.48\linewidth,width=.48\linewidth]{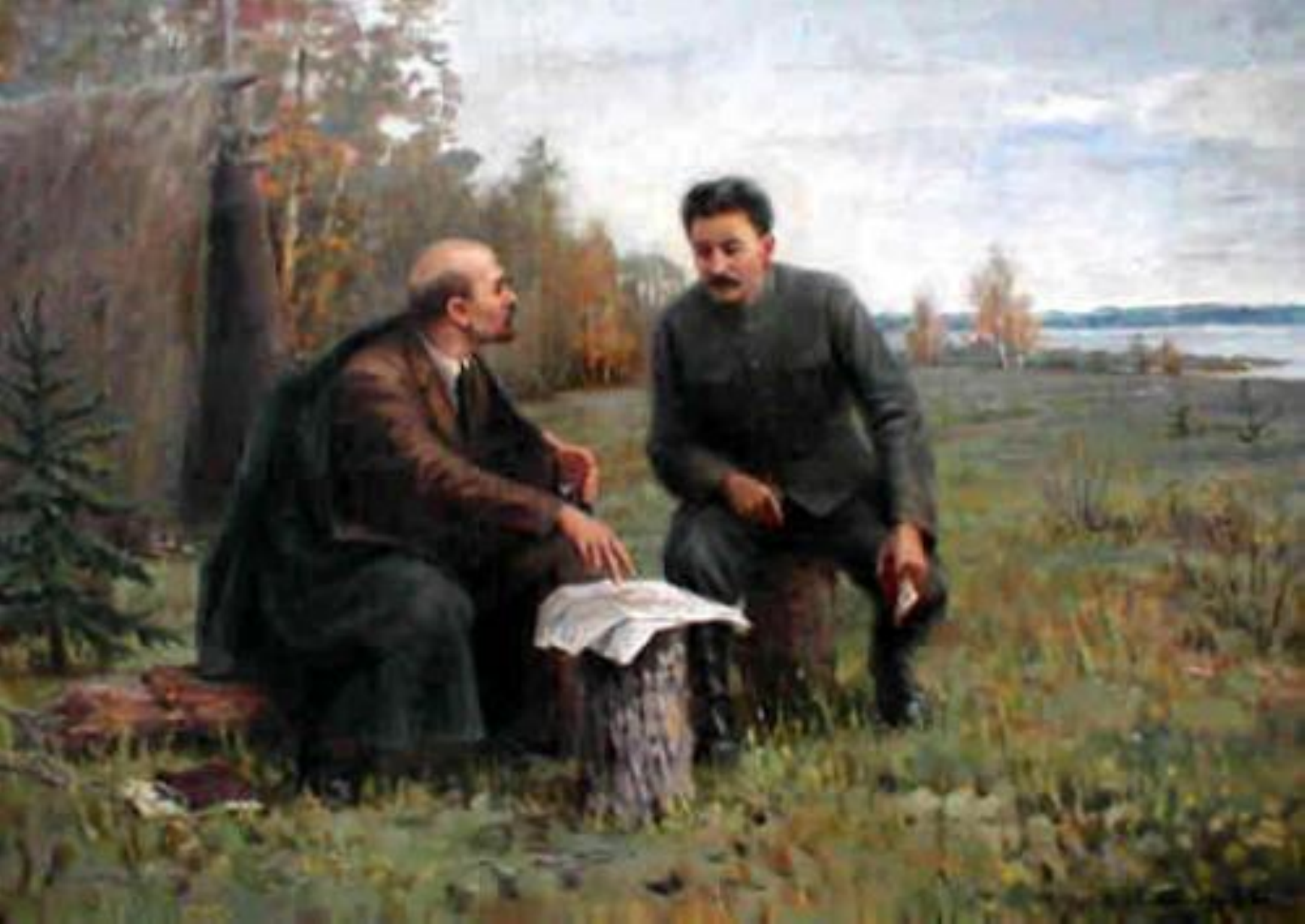}}\\
    {\includegraphics[height=.48\linewidth,width=.48\linewidth]{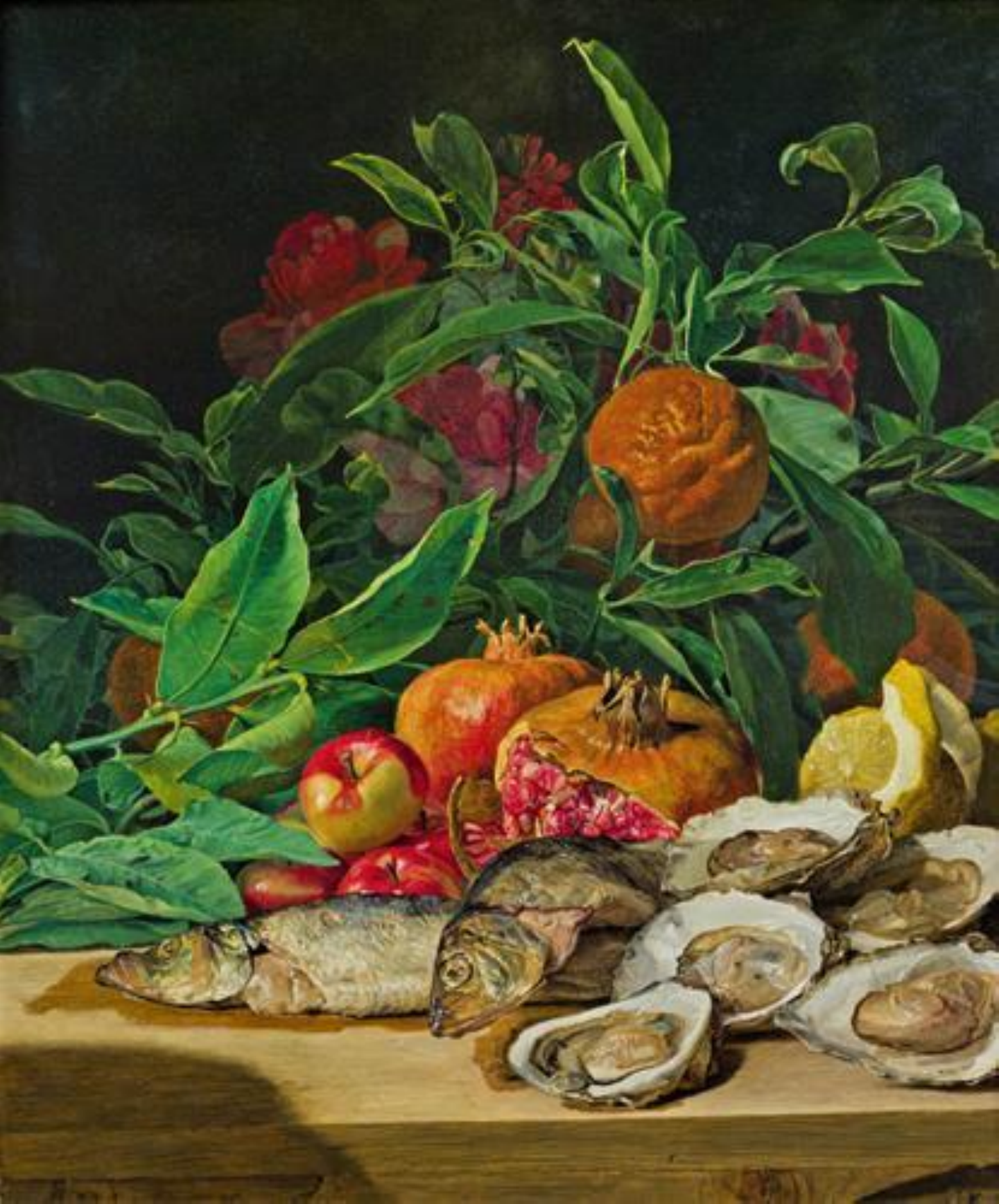}}
    {\includegraphics[height=.48\linewidth,width=.48\linewidth]{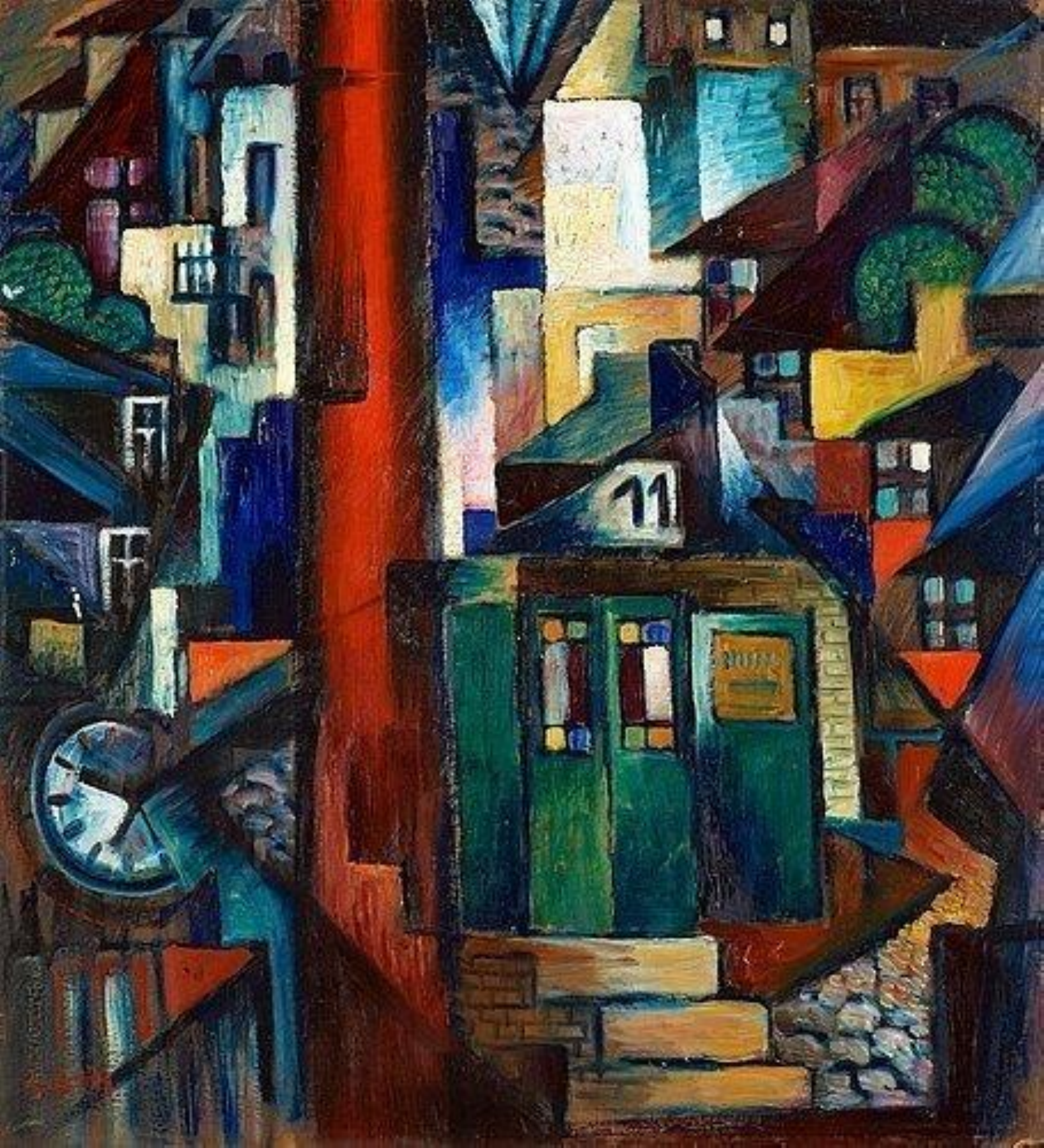}}
    \caption{\small Sample images of the Art category of {\sc Savoias} (styles Suprematism, Socialist Realism, Biedermeier, and Cubo-futurism~\cite{westlake2016detecting}) shown in   
    increasing order of ground-truth visual complexity scores (10, 55, 71, and 83 out of 100).}
    \vspace*{-0.5cm}

\end{figure}
  \vspace{-0.75cm}

\section{Introduction}
Visual complexity is a broad concept with decades of basic and applied research in a variety of areas such as psychophysics and cognitive psychology, product design, marketing, and computing. 
Various definitions can be found in the literature.  One definition relates visual complexity to the level of intricacy and details in an image or the level of difficulty to describe an image~\cite{heaps1999similarity, snodgrass1980standardized}, which can be measured by extracting features from the image~\cite{cardaci2009fuzzy}.  
Another definition of visual complexity is based on the level of visual clutter and the amount of information conveyed in the image, which makes the study of visual complexity related to image compression and information theory~\cite{rosenholtz2007measuring}.

Analysis of visual complexity facilitates assessment of how a human observer may perceive the image, for example, its aesthetic appeal, and how the  observer may interact with it. It has applications in a variety of research and technology fields. Computer vision, graphics, art, interior design, Web design, marketing and advertising are a few examples of the fields that can benefit from the analysis and computation of visual complexity.
Visual complexity is connected to a variety of problems in the computer vision field. 
Visually complex images would need more sophisticated algorithms to perform visual search in the image or create a caption for it~\cite{lin2014microsoft}. It is more challenging to detect or segment objects in an occluded scene where the object might be partially masked. Moreover, 
a visual question answering algorithm could benefit from analysis of the complexity of image regions -- a visually complex region is likely to need more algorithm-generated questions and answers. In addition to computer vision problems, understanding visual complexity of images is beneficial in the context of computer graphics. For example, the more complex a 3D scene is, the more time would take for an algorithm to render it ~\cite{RamanarayananBaFeWa08}. 

Visual complexity is a dominant factor in determining the pleasingness of a stimulus and is known to be related to aesthetic preference for artistic works~\cite{forsythe2011predicting}. 
In ``optimal arousal theory,'' it is shown that the relation between visual complexity and preference follows an inverted U-curve in which images with intermediate level of visual complexity are most appealing~\cite{Berlyne73}. 
In addition, it has been studied that the perception of appeal from a Web design has connection with the visual complexity of the Website~\cite{reinecke2013predicting}, and thus understanding visual complexity can lead to a better subjective experience for users~\cite{krishen2008perceived}. 
Furthermore, advertisement is the first step in eventually persuading consumers to adopt the brand; therefore, advertisers are always looking for ways to motivate consumer engagement with their work. Visual complexity, as one of the forms of complexity in an advertisement can increase arousal by increasing the cognitive demands in the customer~\cite{Huhmann03}.

For quantifying the visual complexity, various factors should be taken into account. The number of objects, colorfulness, edge density, luminance, patterns, and mirror symmetry are some of the examples. However, depending on the type of image, e.g., abstract patterns versus real-world scenes, the contribution of these factors are different.
Due to the wide range of applications for the analysis of visual complexity, the study of visual complexity requires adequate amount of data for different types of images. In addition, the visual complexity  ground truth for the data should be obtained in consistent set of experiments to facilitate the development of new objective algorithms for the analysis of visual complexity.

Studies of visual complexity has a long history in the literature~\cite{Berlyne73,birkhoff1933aesthetic,chipman1977complexity,eysenck1941empirical} and in the past few years, various datasets have been developed~\cite{chipman1977complexity,corchs2016predicting,gartus2017predicting,nadal2010visual}.  These datasets, however, are insufficient for the applications mentioned above.
They are either very small or lack diversity (both number and type of categories, as well as diversity of image content and appearance within each category). In addition, the methodology to obtain the ground truth for these datasets is not consistent.

In this paper, we make three contributions to the study of visual complexity in images:

\begin{enumerate}
    \item We introduce {\sc Savoias,} a dataset for the analysis of visual complexity. {\sc Savoias} covers a variety of topics and provides a sufficient number of images per topic, therefore improving diversity and scale of publicly available datasets. Specifically, {\sc Savoias} consists of seven diverse categories and a total of 1,420 images.
 {\sc Savoias} is an acronym for {\bf S}cenes, {\bf A}dvertisement, {\bf V}isualization and infographics, {\bf O}bjects, {\bf I}nterior design, {\bf A}rt, and {\bf S}uprematism (a category of art). 
    \vspace*{-0.2cm}
    \item In order to minimize the potential bias from individual ground-truth contributors and limited rating scales, we obtained the ground-truth labels 
using a forced-choice crowdsourcing methodology on a [0,100] rating scale.  Labels were obtained from 1,687 contributors on more than 37,000 pairs of images. The pairwise scores were then converted to absolute scores using the Bradley-Terry algorithm and matrix completion.
    \vspace*{-0.2cm}
    \item We explore the applicability of five state-of-the-art algorithms on the different categories of our dataset and report the gap between their performance and the ground truth. Our results highlight the need for the development of new, potentially category-specific techniques.
\end{enumerate}

\section{Related Work}
\begin{table*}[t]
\renewcommand{\tabcolsep}{1mm}
  \small
  \centering
    \caption{\small The datasets previously used in visual complexity studies, as well as our proposed dataset. A ``1-step'' groundtruthing process means that the users were asked to directly rate the visual complexity of a single image;
    ``Shared'' means the dataset may be shared with other researchers upon request. 
    \vspace*{-0.5cm}
    }
    \newcolumntype{P}[1]{>{\centering\arraybackslash}p{#1}}

  \begin{tabular}{l||r|l|l|l|l}
    \hline
      &  & &Ground-truth &Vis. Comp. & Open  \\
     Reference & \multicolumn{1}{l|}{\# Images}  & Application Category & Process &scale & Source\\
     \hline \hline
     Gartus~{\it et al.}~\cite{gartus2013small} & 912 & Black and white $8\times8$ abstract patterns &1-step &5-point  & Shared \\\hline
      Nadal~{\it et al.}~\cite{nadal2010visual} & 120 & Abstract \& representational (artistic \& non-artistic) &1-step& 3-point  & No \\\hline
   
     Olivia~{\it et al.}~\cite{olivia2004identifying} & 100 &  Indoor scenes &   3-step &8-point & No \\\hline
     
     Miniukovich~{\it et al.}~\cite{MiniukovichAn14} & 140 &  Webpage &1-step& 5-point  & No \\\hline
     
     Fan~{\it et al.}~\cite{FanLYZ17} & 40 & Chinese ink painting (abstract and representational) &1-step& 7-point  & No \\\hline
     
      
     Schnur~{\it et al.}~\cite{SchnurBeCo18}& 9 & Web maps&1-step &5-point  & No\\\hline 
    Corch~{\it et al.}~\cite{corchs2016predicting} & 98 & Real-world scenes &1-step& [0-100]  & Yes  \\\hline
     
    Corch~{\it et al.}~\cite{corchs2016predicting}& 122 &  Textures& 1-step&[0-100]  & Shared \\ \hline
    
    {\bf \multirow{2}{*}{Ours}} & {\bf \multirow{2}{*}{1,420}} & {\bf Scenes, advertisement, visualizations, objects, } & {\bf Pairwise} &{\bf \multirow{2}{*}{[0-100]}}& {\bf \multirow{2}{*}{Yes} }\\
    &&{\bf interior design, art, Suprematism}&{\bf comparison}&  &\\
    \hline
  \end{tabular}
  \label{table:datasets}
\end{table*}

Visual complexity has been previously studied extensively in the literature of psychophysics, cognitive science, and more recently in computer vision. 
While the temporal dimension of complexity is also an interesting topic of research~\cite{ cardaci2009attentional, marin2016effects,palumbo2014examining}; in this work, we focus on the spatial dimension of visual complexity and algorithmic approaches to quantify it.

{\bf Applications of Visual Complexity.}
Existence of a linear relation between visual complexity and aesthetic beauty has been previously stated in the literature~\cite{eysenck1941empirical,reinecke2013predicting}. Jacobsen and H\"{o}fel~\cite{jacobsen2001aesthetics} argued that symmetry and visual complexity are two predictors for aesthetic judgment of beauty. Birkhoff came up with the formula $M=O/C$, where $M$ refers to aesthetic measurement, $O$ refers to aesthetic order, and $C$ refers to complexity. Based on his formula, beauty decreases with increase in complexity~\cite{birkhoff1933aesthetic}.

Furthermore, studies show that in hedonic shopping experiences, the higher perceived complexity of mall interiors yields a higher satisfaction score while in utilitarian shopping experiences, people prefer lower visual complexity~\cite{haytko2004s}. For online shopping, the perceived visual complexity has been shown to negatively influence individuals’ satisfaction~\cite{SohnSeMo17}. 

Visual impression of advertisement images plays a crucial role in engaging visitors. These first impressions further affect the mid- and long-term human behavior. Pieters~{\it et al.}~\cite{PietersWeBa10} have considered two different types of complexity for ads: feature complexity (depends on the visual features), and design complexity (depends on the creative design) both of which indicate perceptual complexity. 
They argued that feature complexity hurts attention to the brand, whereas design complexity can improve the consumer's attention.

{\bf Algorithms.} 
In one of the early works on visual complexity, Chipman~{\it et al.}\ explained the importance of two factors in the analysis of visual complexity, specifically, quantitative factor (related to amount of elements) which has positive correlation with visual complexity and structural element ( determined by different forms of structural organization, but mostly by symmetry) which has negative correlation with visual complexity~\cite{chipman1977complexity,chipman1979influence}.  
In a similar approach, a more recent study explored the impact of these two factors on the visual complexity of more complex abstract patterns~\cite{gartus2013small,gartus2017predicting}. 

The applicability of this method was further demonstrated by Nadal {\it et al.} on four categories: abstract artistic, abstract non-artistic, representational artistic, and representational non-artistic. The images used in their experiment contained 120 stimuli equally divided into three complexity
levels: low, intermediate, and high~\cite{nadal2010visual}.

Oliva~{\it et al.} studied the impact of task constraint on representation of visual complexity by creating a visual complexity dataset of 100 indoor images. They argued that although the contribution of the perceptual dimensions are affected by task constraints, visual complexity can still be represented by perceptual dimensions such as quantity of objects, clutter, openness, symmetry, organization, and variety of colors~\cite{olivia2004identifying}. In addition, five factors encapsulating visual complexity for web design and GUI applications are proposed~\cite{MiniukovichAn14}. Similarly, Fan~{\it et al.} expanded this approach to Chinese ink paintings by introducing three new features, namely color richness, stroke thickness, and white spaces~\cite{FanLYZ17}. Furthermore, in a more recent study, linear combination of multiple features such as edge density, compression ratio, and number of objects was proposed. The images used in their experiments consists of real-world scene images and textures. They obtained ground truth on a scale in the range of 0 to 100.

Visual complexity can also be approximated using algorithmic information theory and compression algorithms~\cite{rosenholtz2007measuring}. In this case, the visual complexity is defined as the resulting file size when a compression algorithm such as JPEG or ZIP is applied on the given image. The stimuli used in these experiments were geographical maps and synthetic images of objects in sparse arrangements. 
In another study, visual complexity of three different online map providers (Google Maps, Bing Maps, and OpenStreetMaps) was explored with the objective of better understanding design decisions for Web maps. 
Their results implied that clutter, measured by feature congestion~\cite{rosenholtz2007measuring}, is more important in perceived complexity than diversity of symbology~\cite{SchnurBeCo18}.

The characteristics of the datasets mentioned in this section are summarized in Table~\ref{table:datasets}. It shows differences in scales and image collection methods, as well as groundtruthing methodologies, and reveals that the diversity and number of samples in these datasets are inadequate for extensive analysis of visual complexity.  It is also worth noting that not all the datasets mentioned above are publicly available or shared among researchers. 

 {\sc Savoias,} is introduced to address the lack of an appropriate dataset. It is a new dataset consisting of seven diverse categories and more than 1,400 sample images. In order to obtain ground truth for our dataset, instead of asking the participants to rate the visual complexity of an image on a continuous scale, we incorporated a pairwise comparison between images to avoid any bias in the rating scale. The pairwise methodology also provides a more fine-grained range of scores compared to the common 3, 5, or 7-scale ratings used in the aforementioned datasets. 

\vspace*{-0.2cm}
\section{Dataset Description}
\begin{table*}[htbp]
\hskip3pt
 \caption{\small Sample images of the {\sc Savoias} dataset with increased visual complexity from left to right in each row.
 }
 \label{image examples}
\begin{center}

\begin{supertabular}{ >{\centering\arraybackslash}m{0.1in}  > {\centering\arraybackslash}m{4cm}  >{\centering\arraybackslash}m{4cm}  >{\centering\arraybackslash}m{4cm}  >{\centering\arraybackslash}m{4cm} }
\begin{turn}{90}{Scenes}\end{turn} & \includegraphics[height=2.8cm, width=4cm]{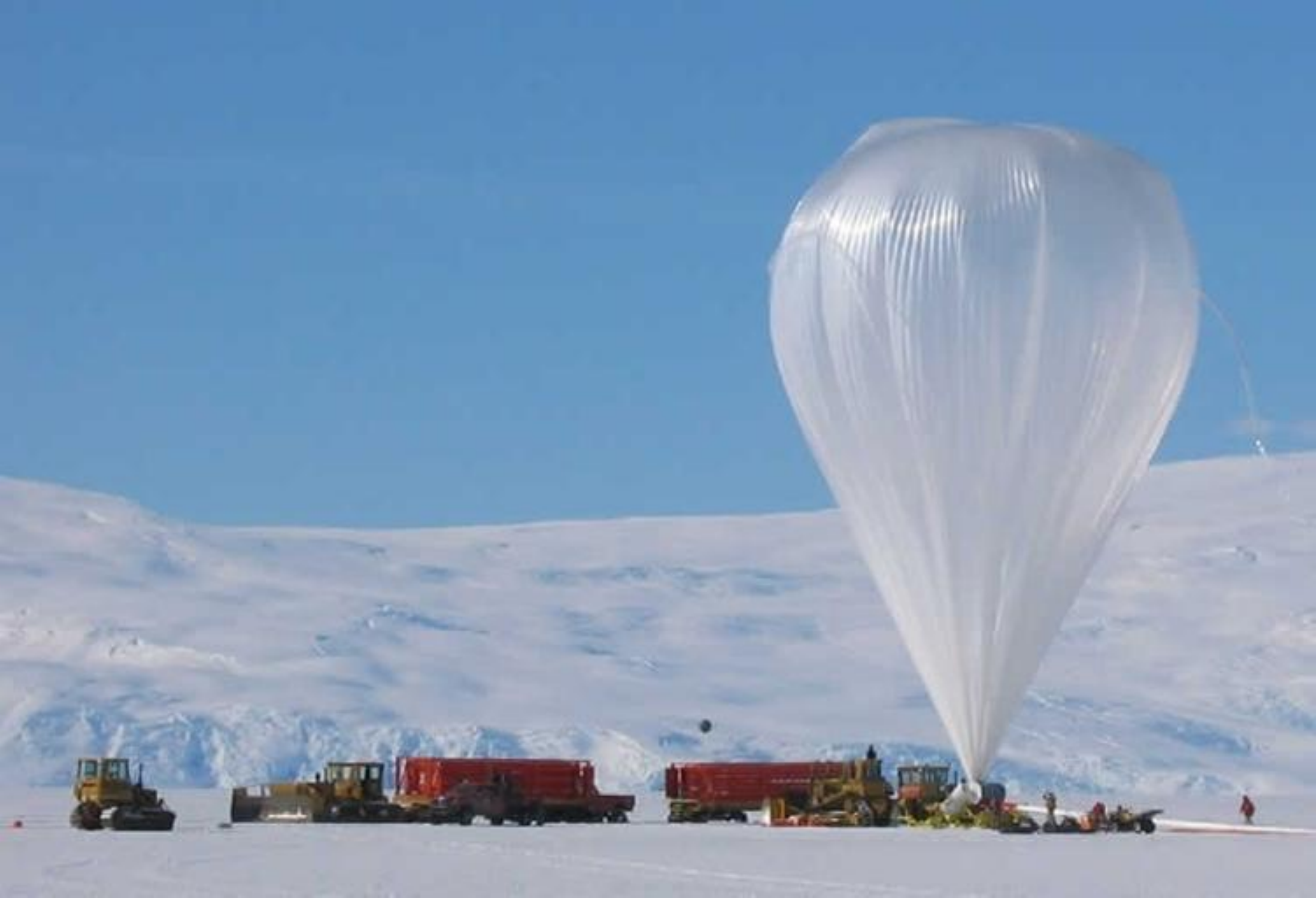} &\includegraphics[height=2.8cm, width=4cm]{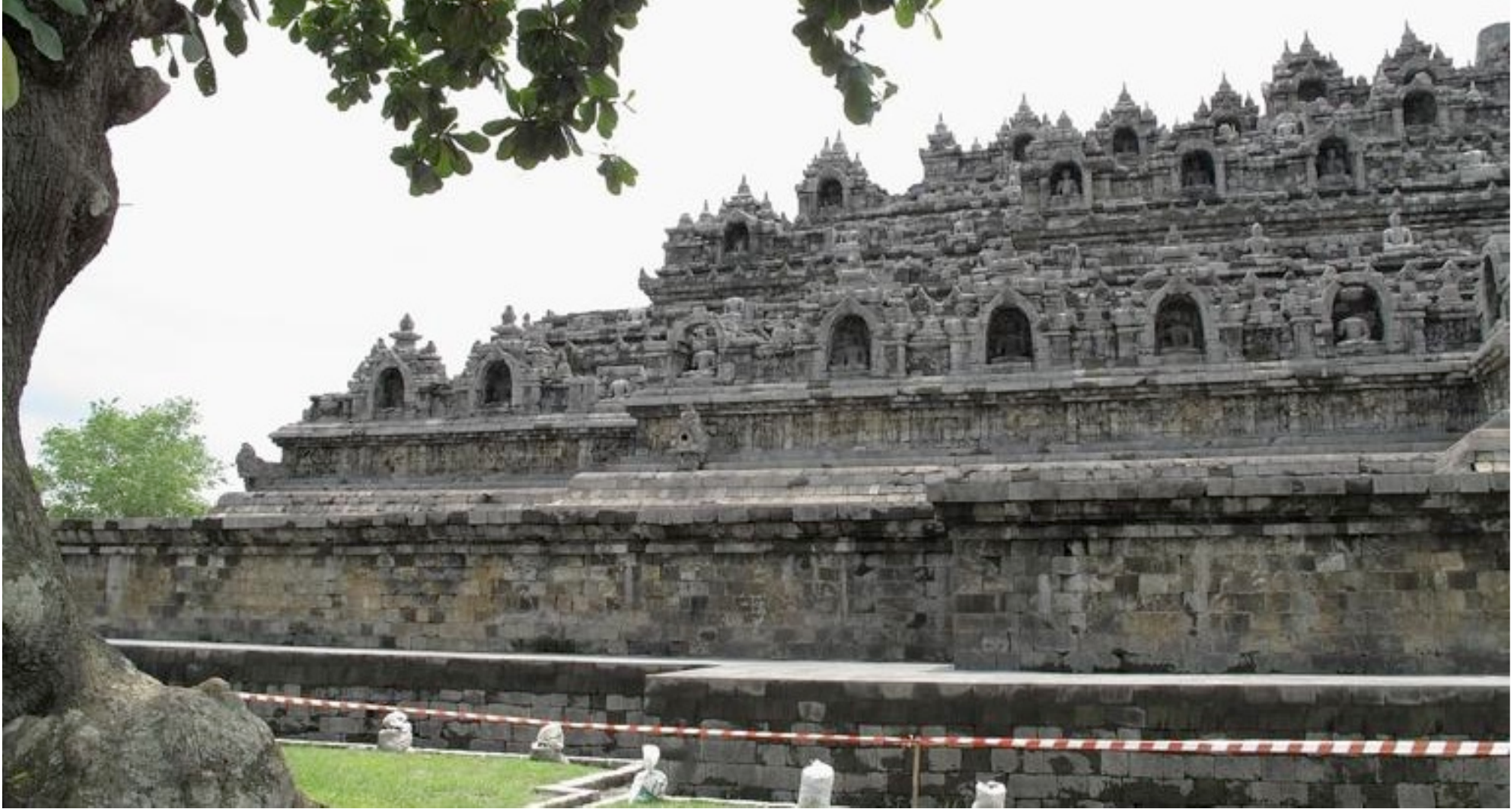} & \includegraphics[height=2.8cm, width=4cm]{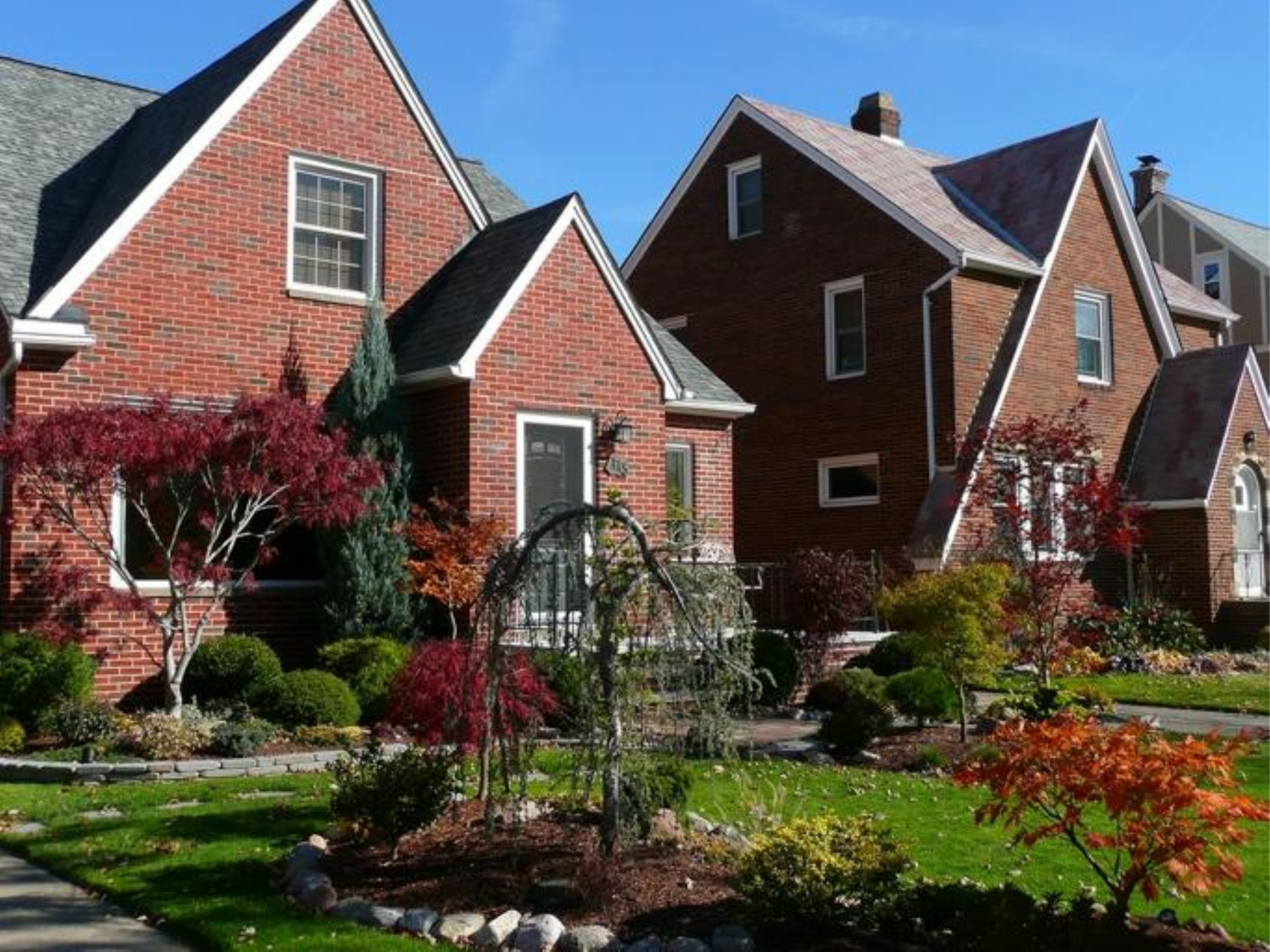}& \includegraphics[height=2.8cm, width=4cm]{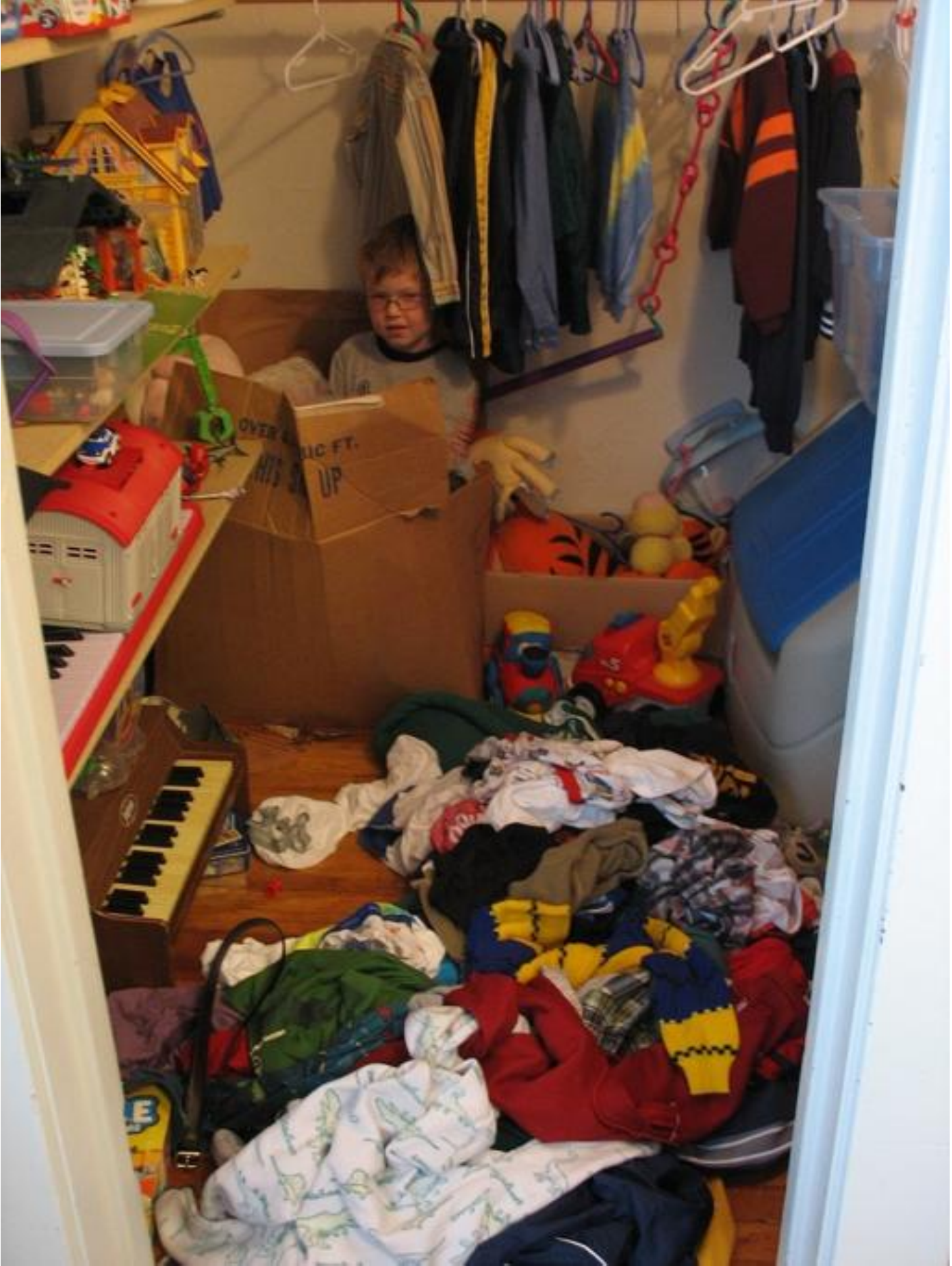} \\
\begin{turn}{90}{Advertisement}\end{turn} & \includegraphics[height=2.8cm, width=4cm]{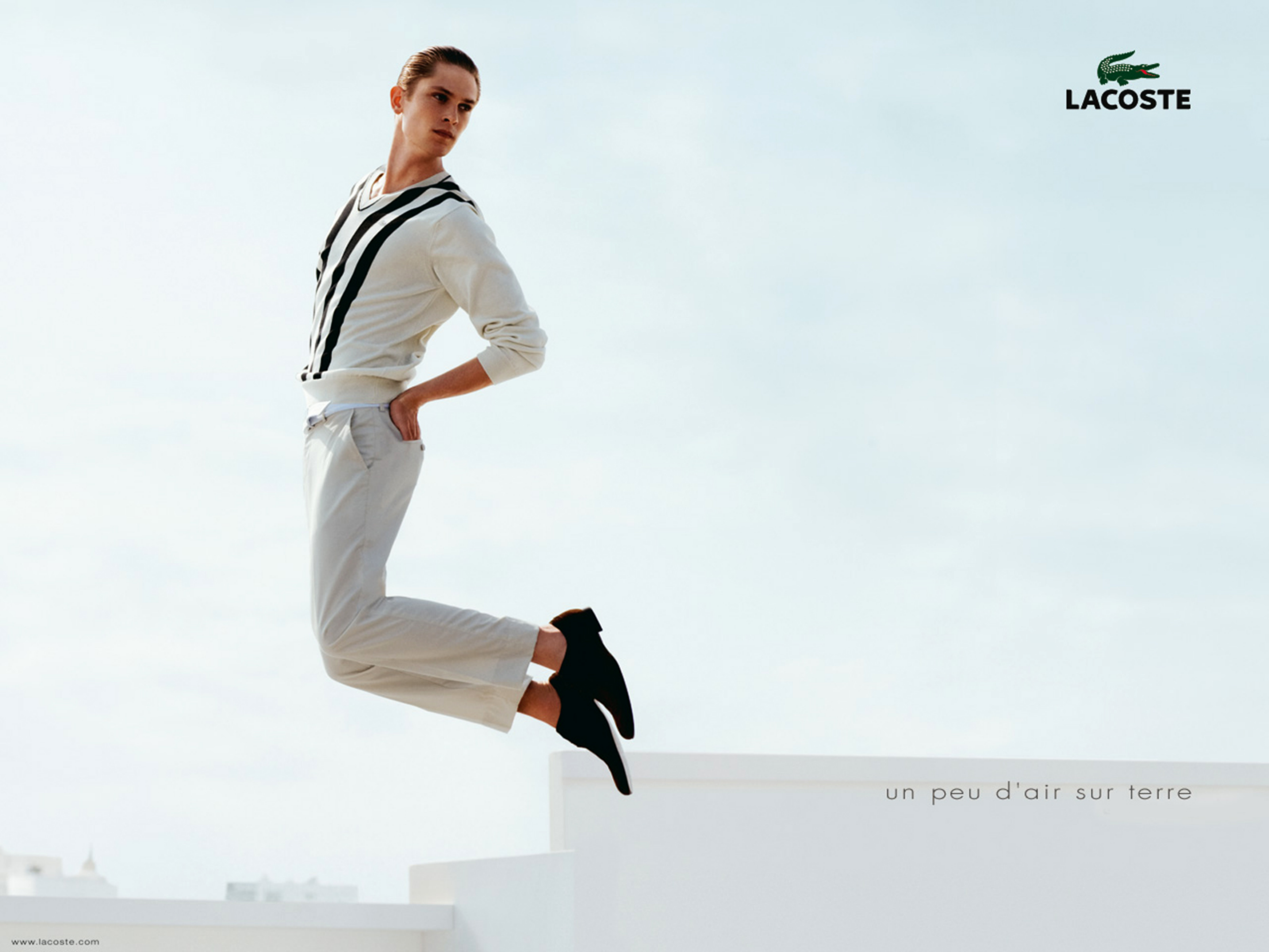} &\includegraphics[height=2.8cm, width=4cm]{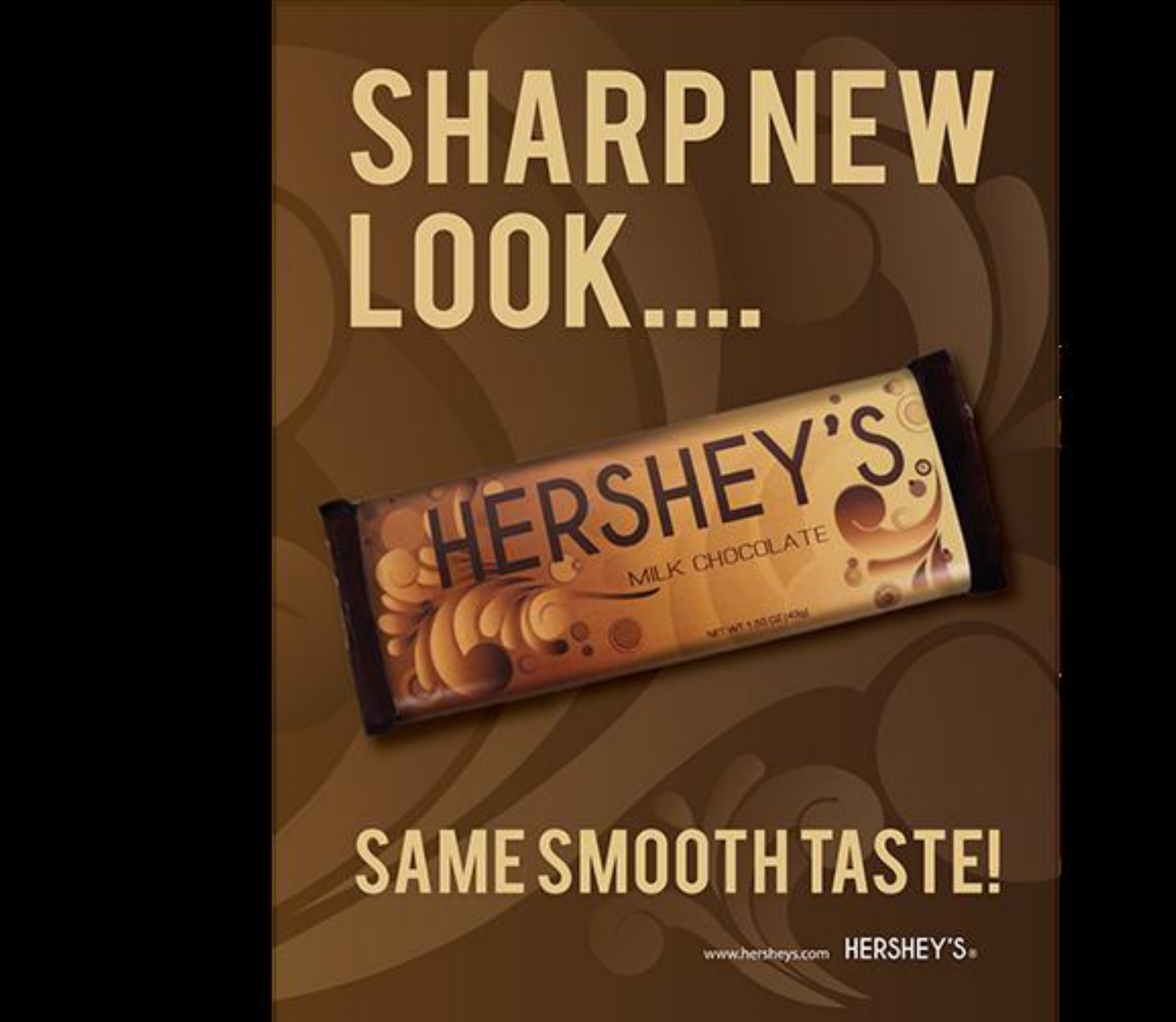} & \includegraphics[height=2.8cm, width=4cm]{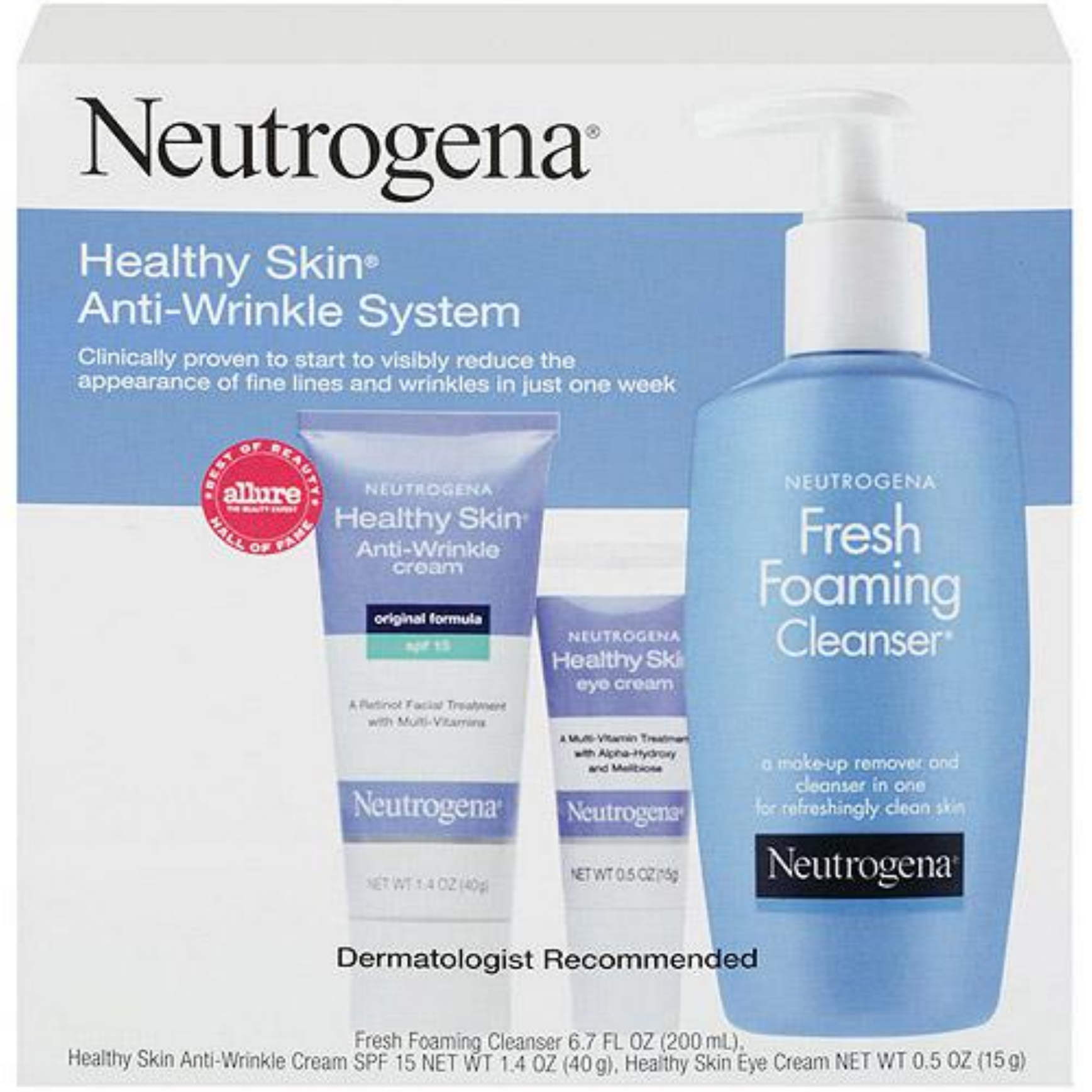}& \includegraphics[height=2.8cm, width=4cm]{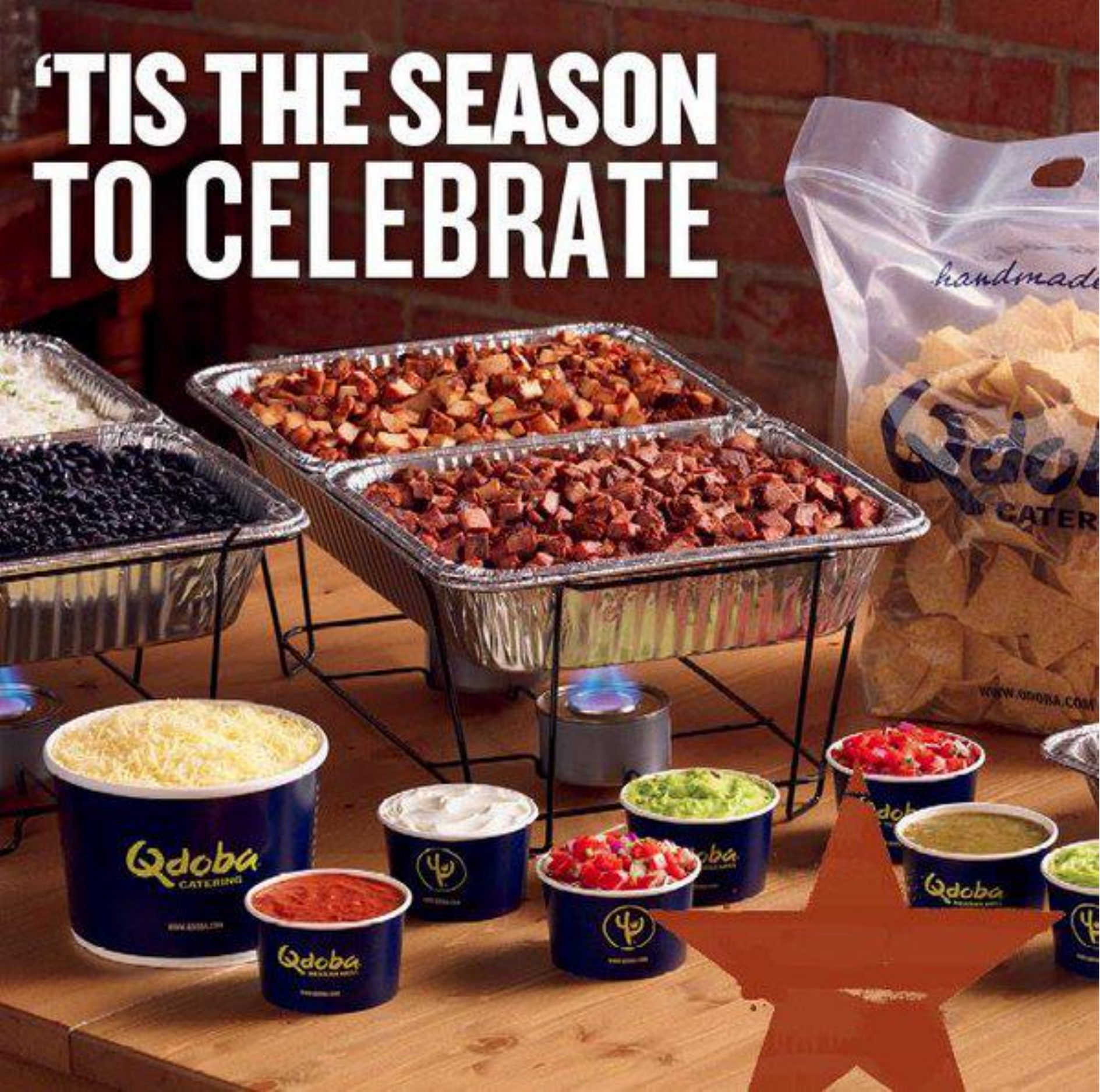} \\
 %
 
\begin{turn}{90}{Visualization}\end{turn} & \includegraphics[height=2.8cm, width=4cm]{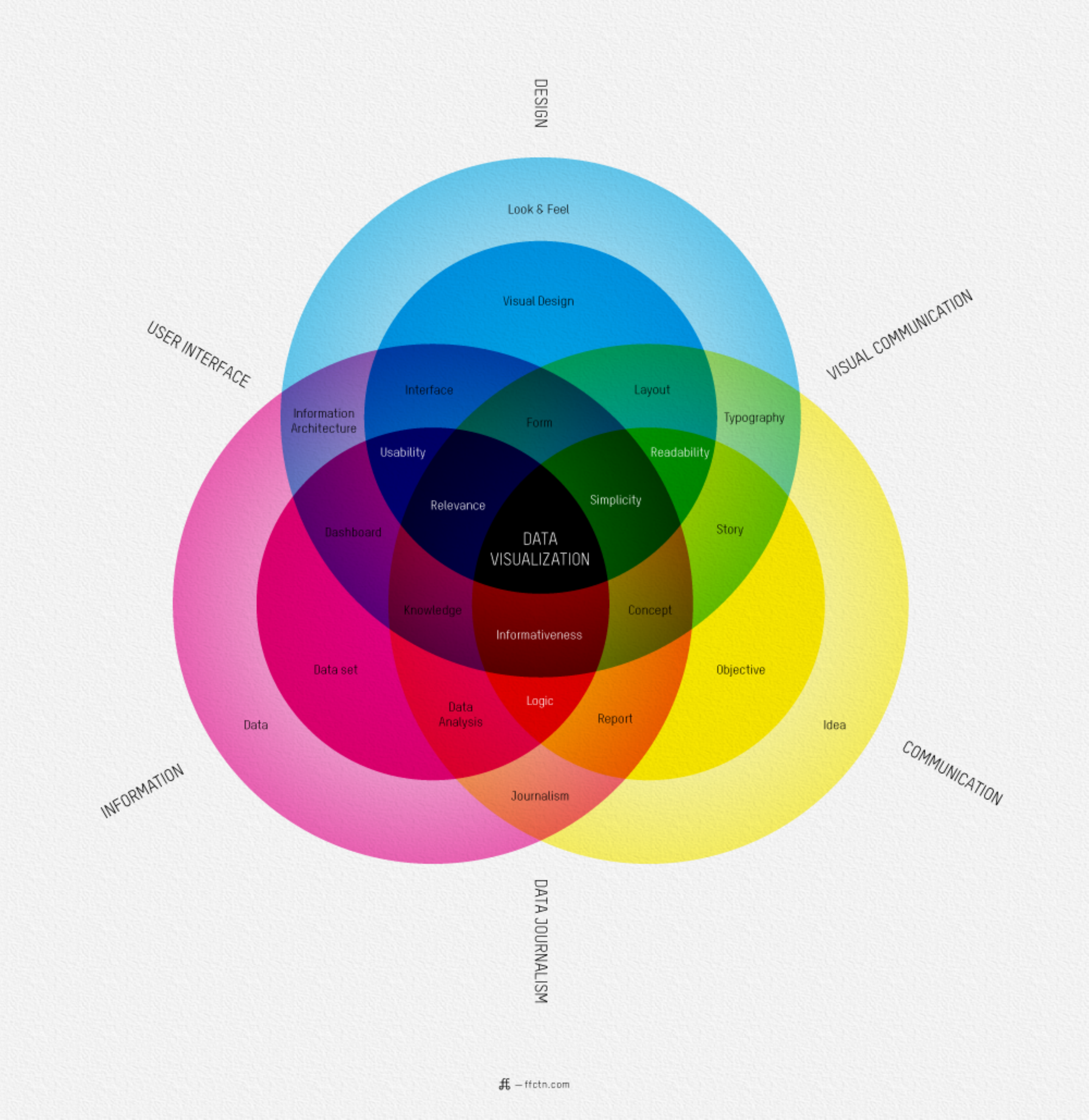} &\includegraphics[height=2.8cm, width=4cm]{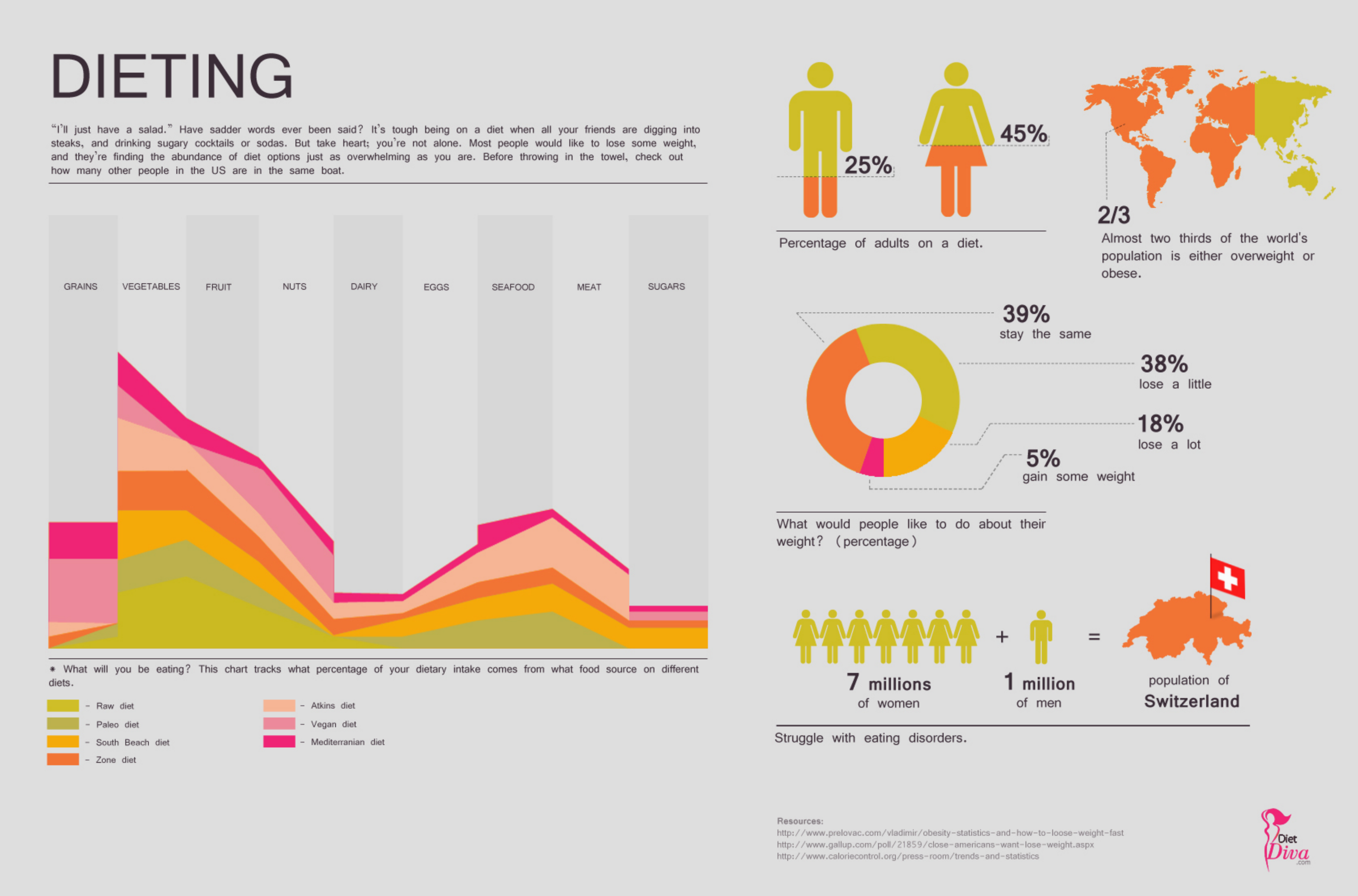} & \includegraphics[height=2.8cm, width=4cm]{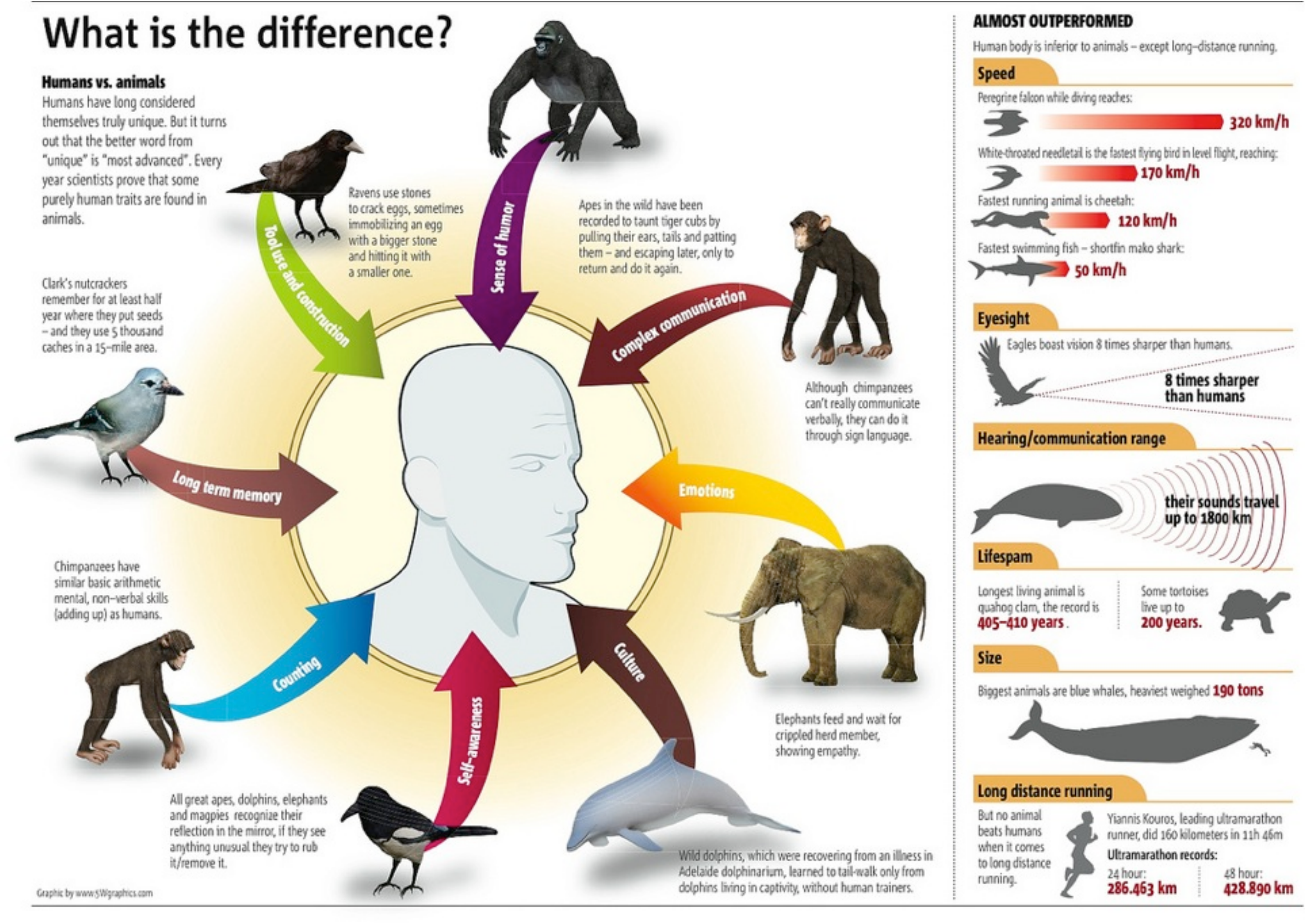}& \includegraphics[height=2.8cm, width=4cm]{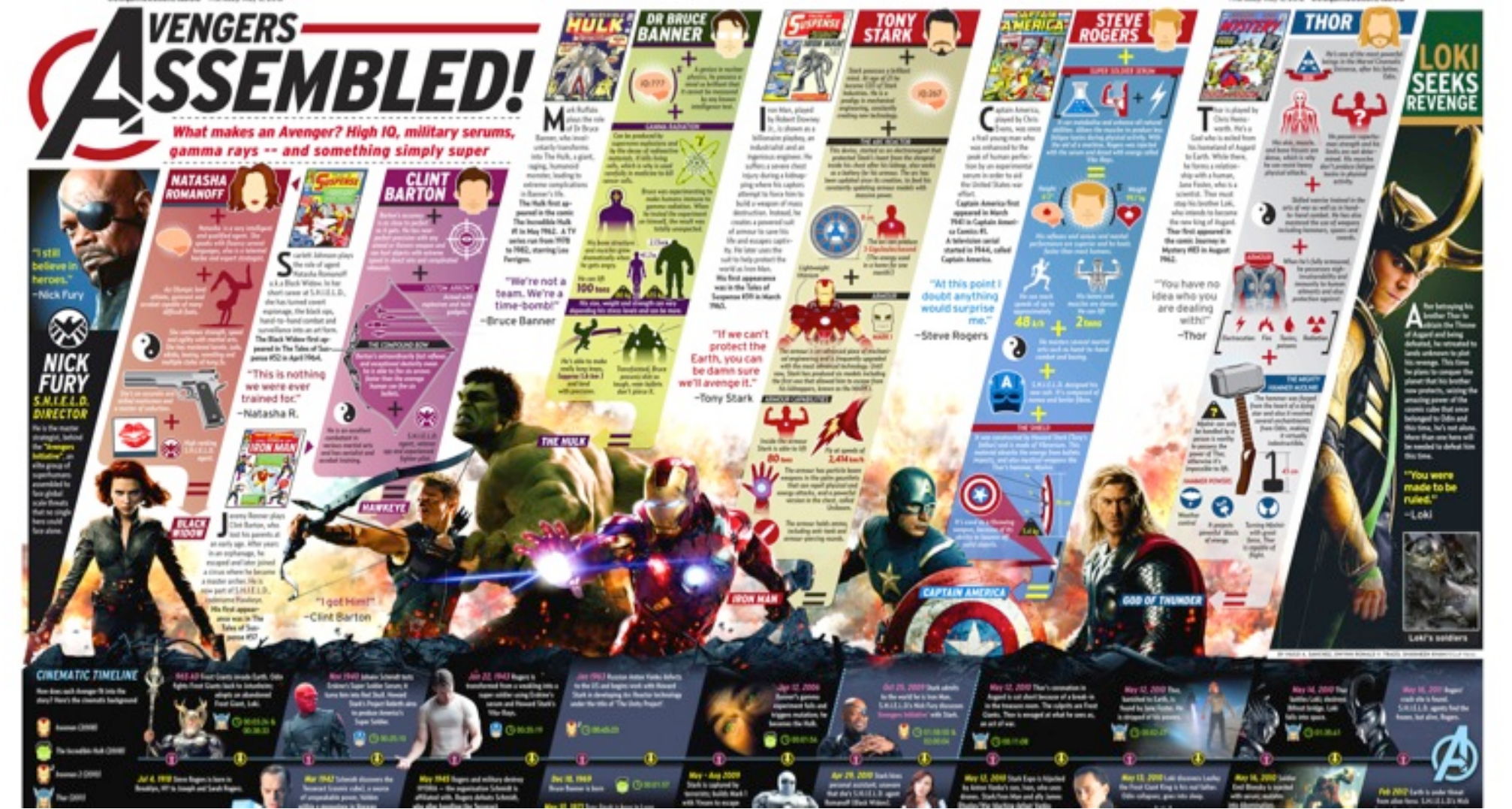} \\

 \begin{turn}{90}{Objects}\end{turn} & \includegraphics[height=2.8cm, width=4cm]{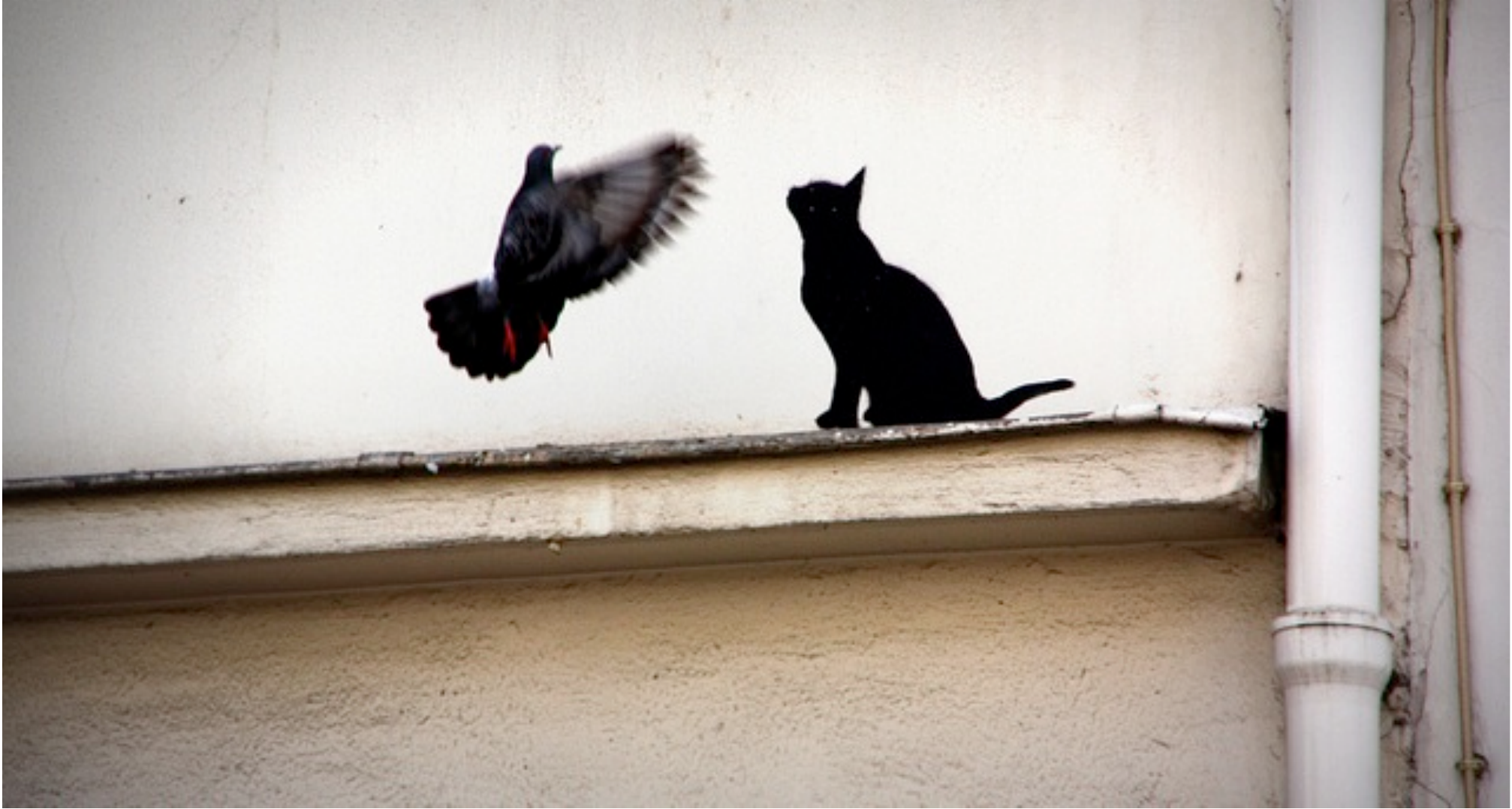} &\includegraphics[height=2.8cm, width=4cm]{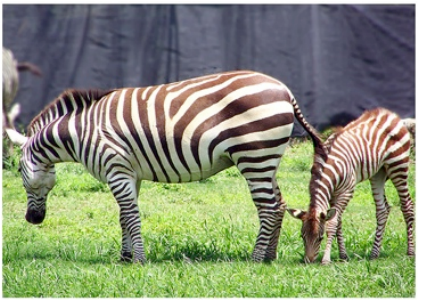} & \includegraphics[height=2.8cm, width=4cm]{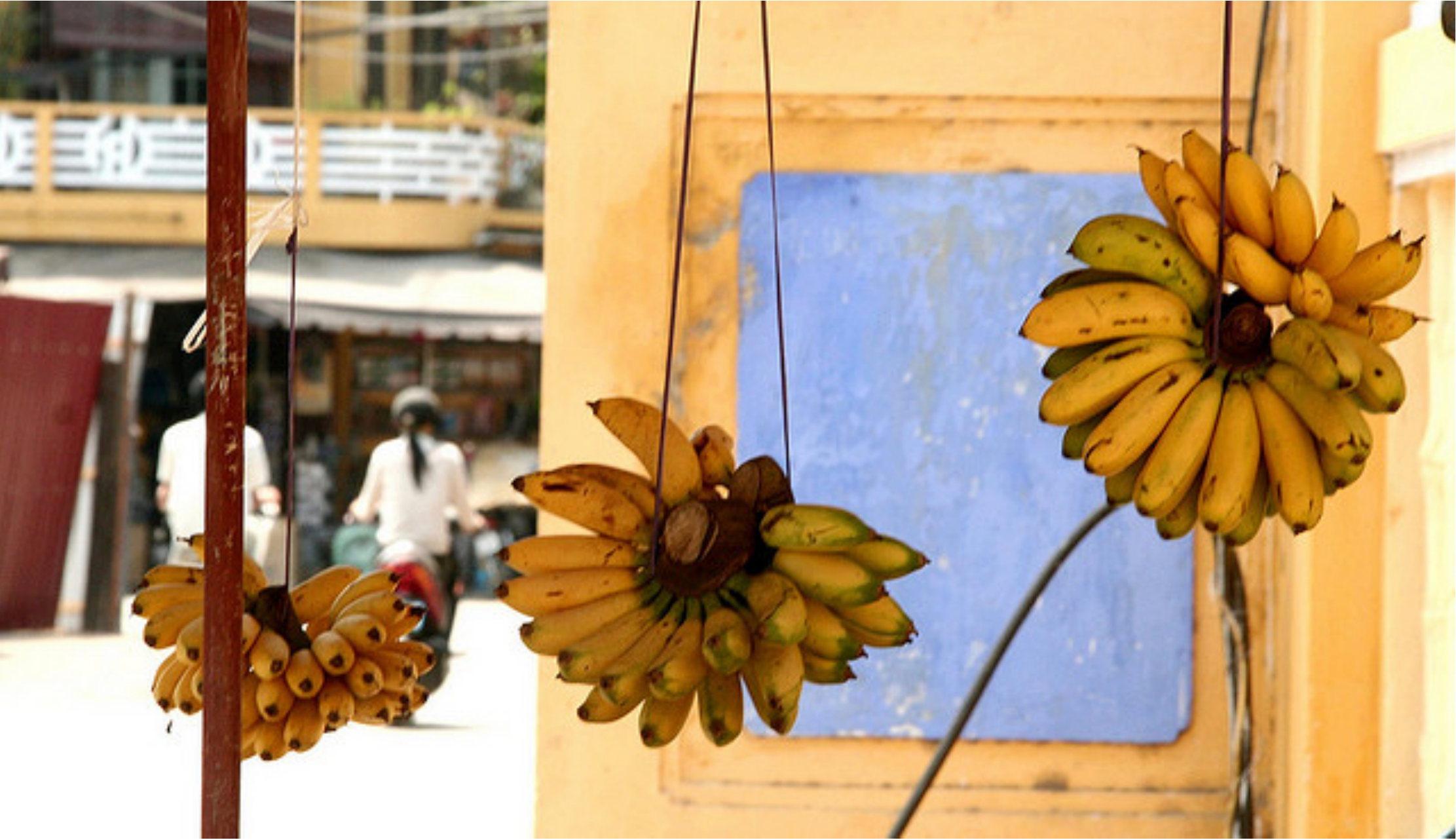}& \includegraphics[height=2.8cm, width=4cm]{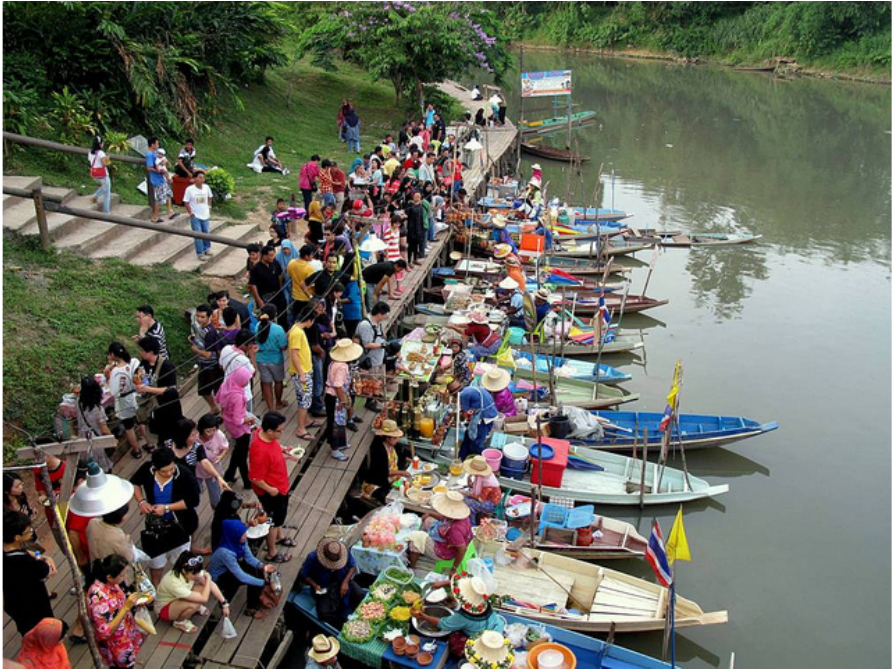} \\ 
  
 \begin{turn}{90}{Interior Design}\end{turn} & \includegraphics[height=2.8cm, width=4cm]{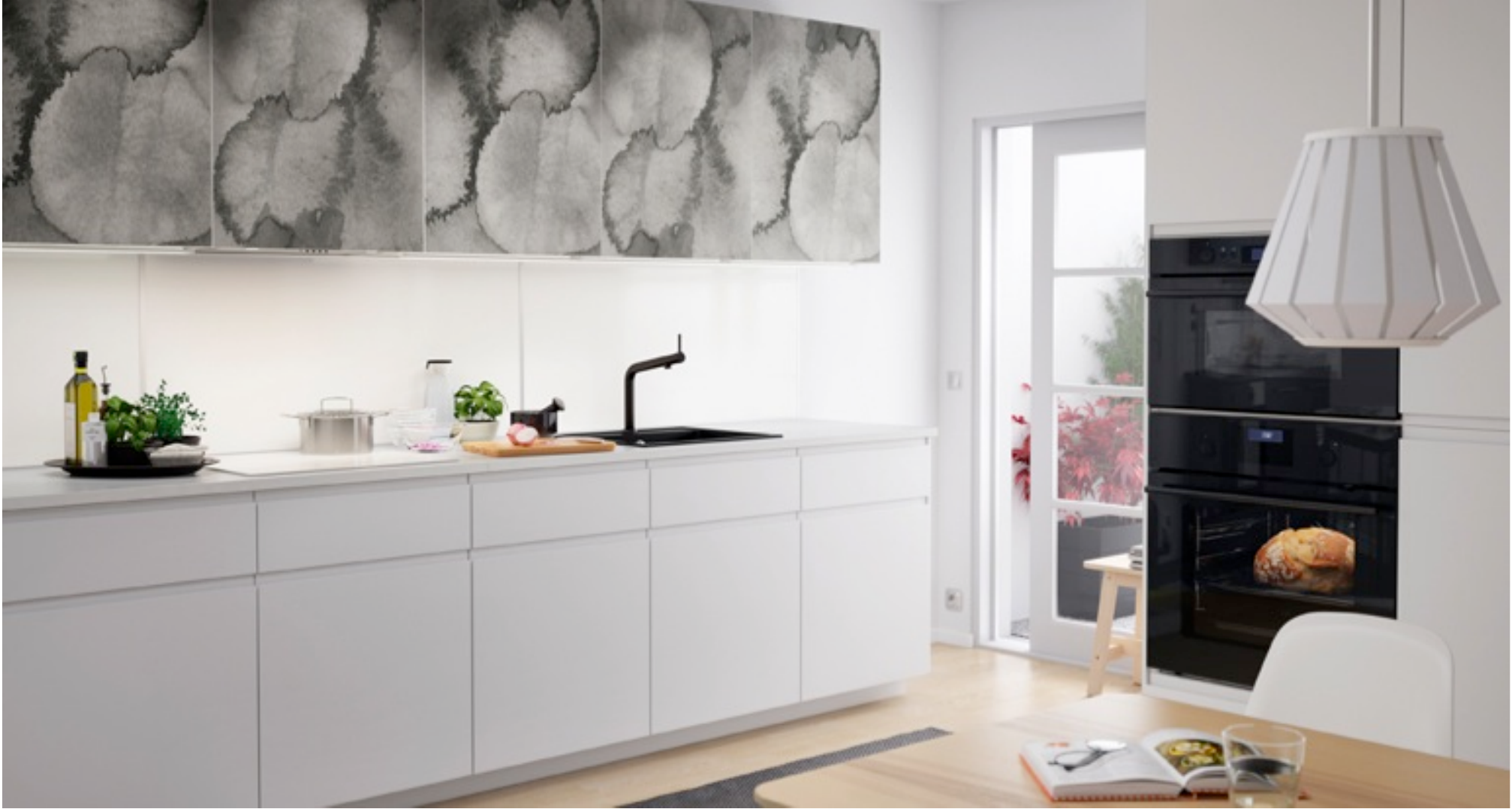} &\includegraphics[height=2.8cm, width=4cm]{23.pdf} & \includegraphics[height=2.8cm, width=4cm]{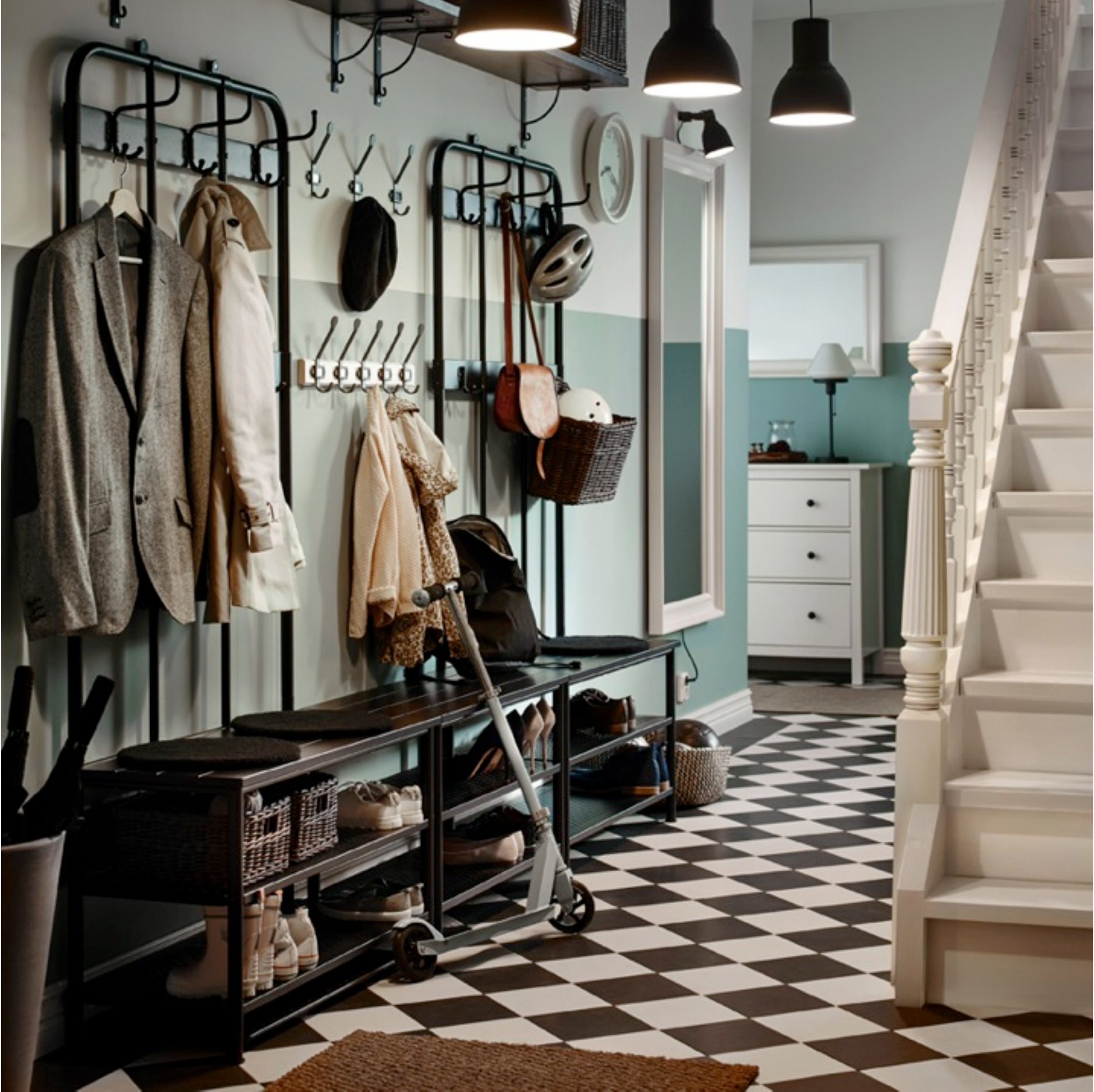}& \includegraphics[height=2.8cm, width=4cm]{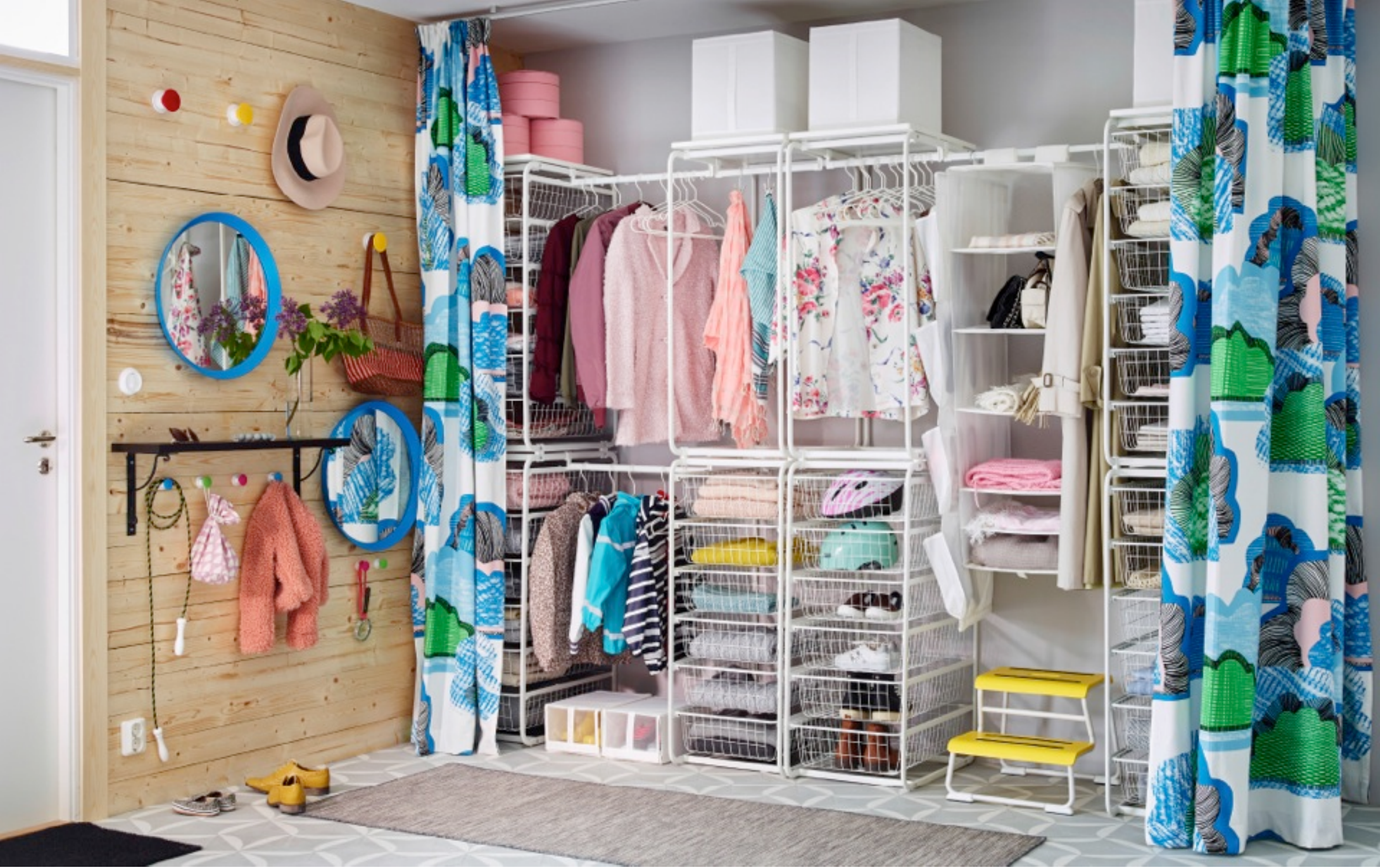} \\
  
\begin{turn}{90}{Art}\end{turn}
& \includegraphics[height=2.8cm, width=4cm]{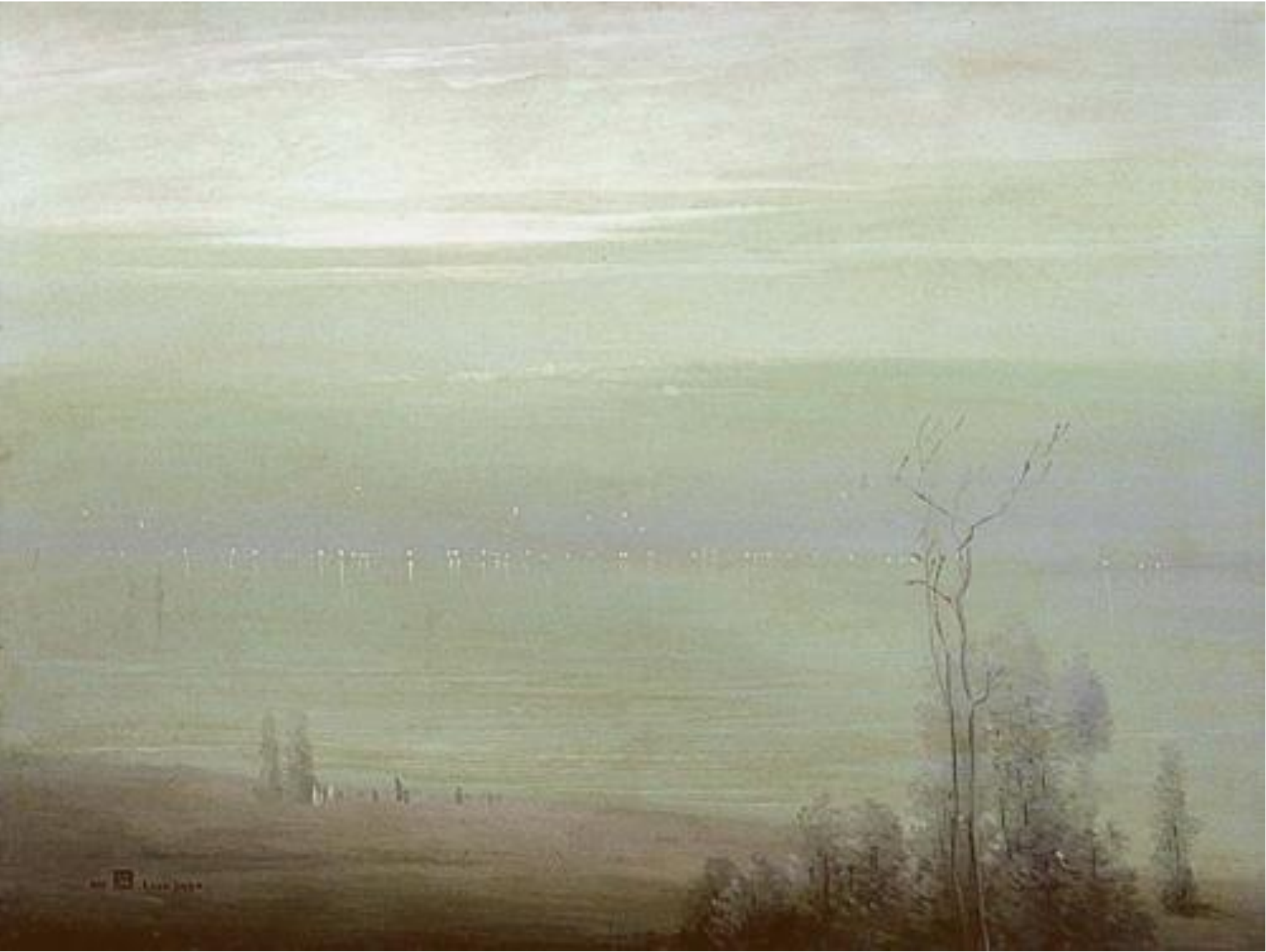} &\includegraphics[height=2.8cm, width=4cm]{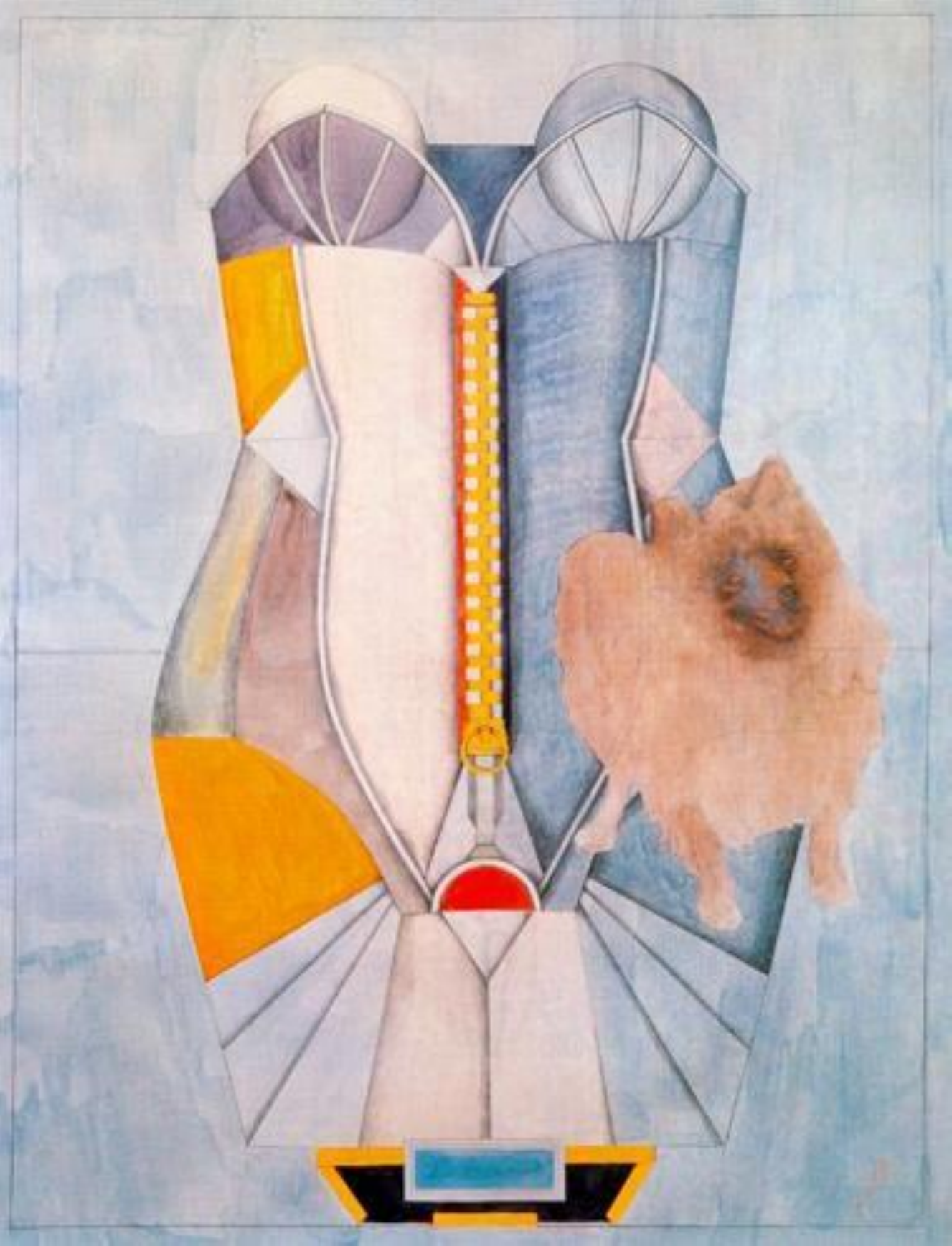} & \includegraphics[height=2.8cm, width=4cm]{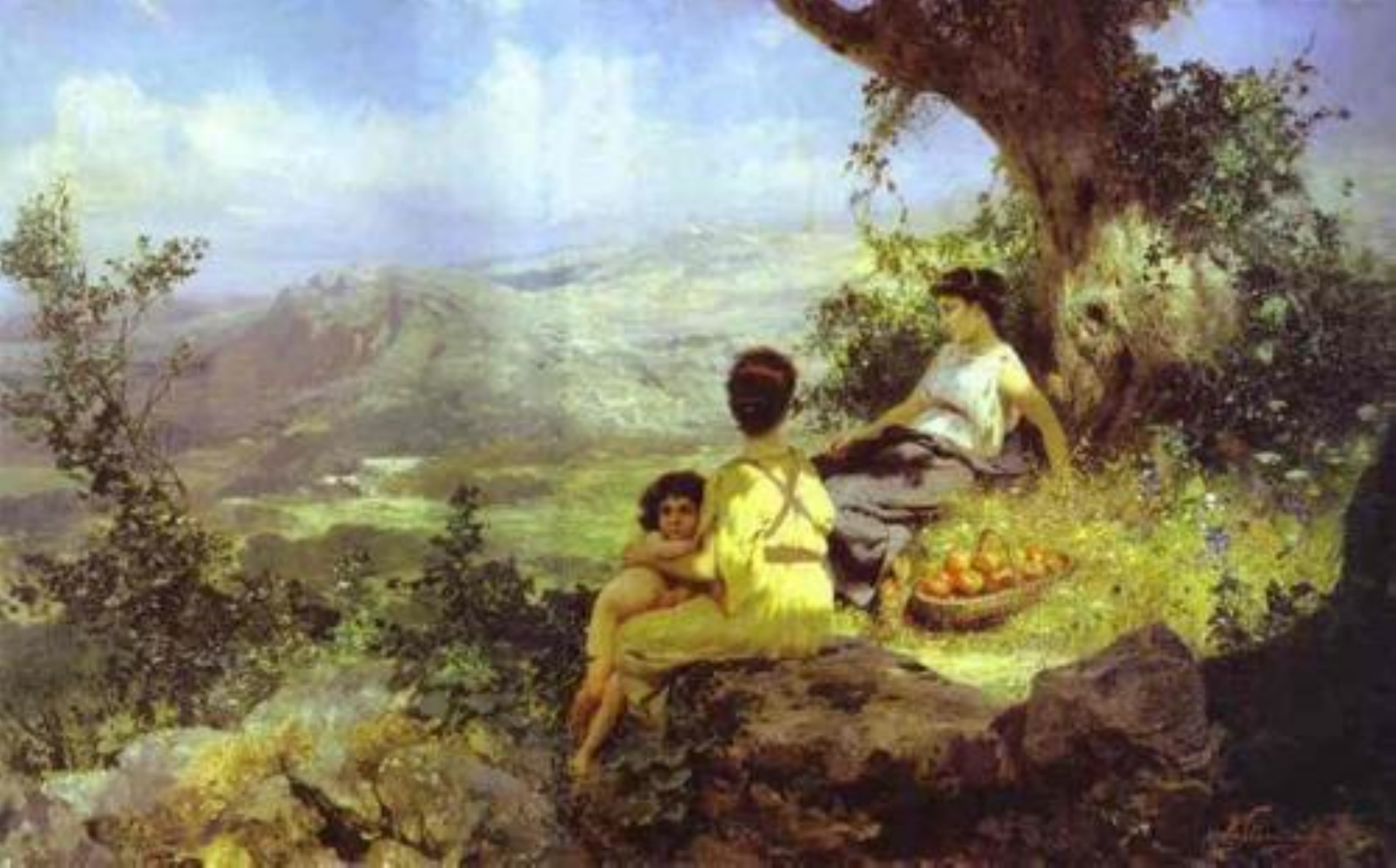}& \includegraphics[height=2.8cm, width=4cm]{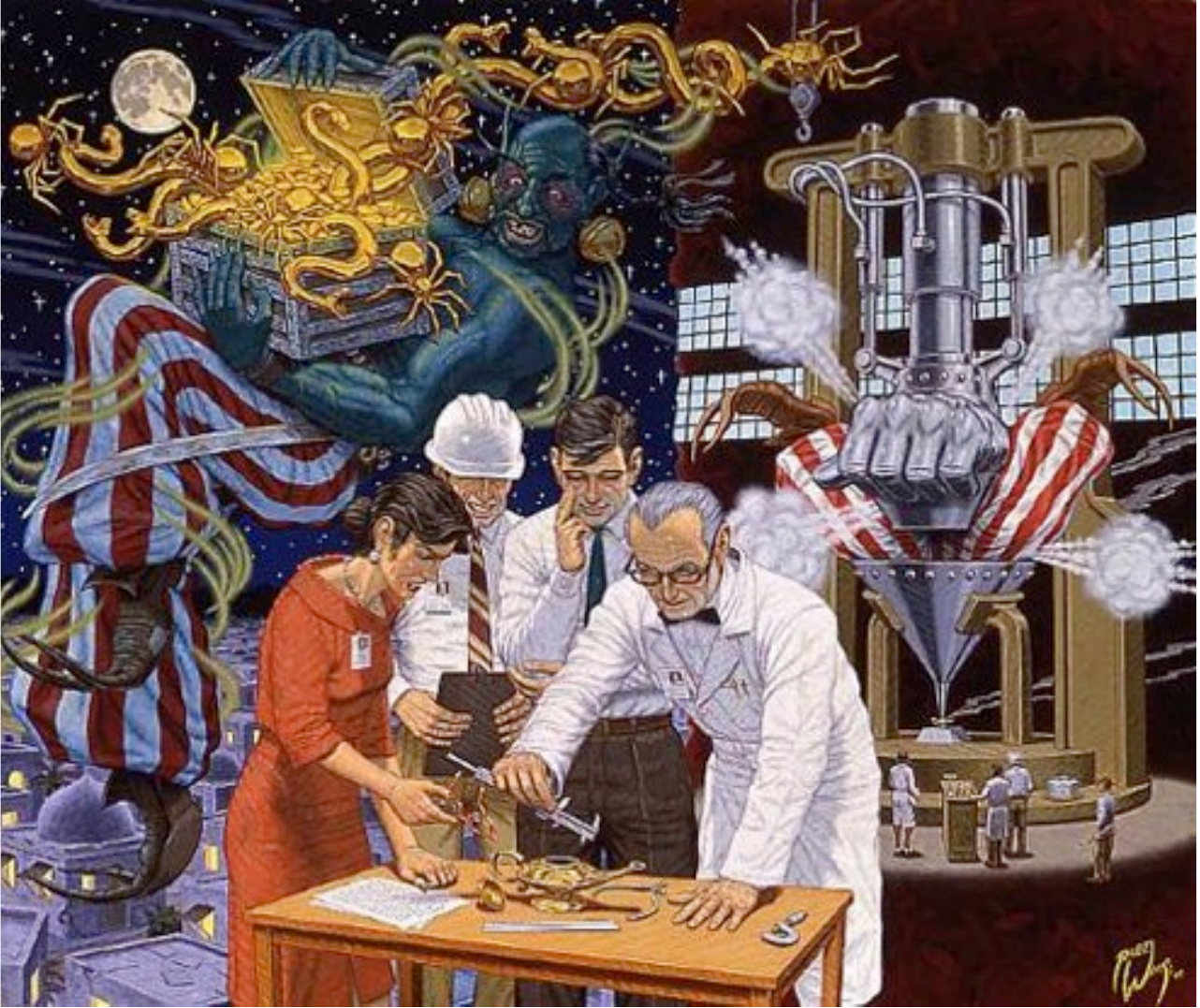} \\
 
\begin{turn}{90}{Suprematism}\end{turn} 
& \includegraphics[height=2.8cm, width=4cm]{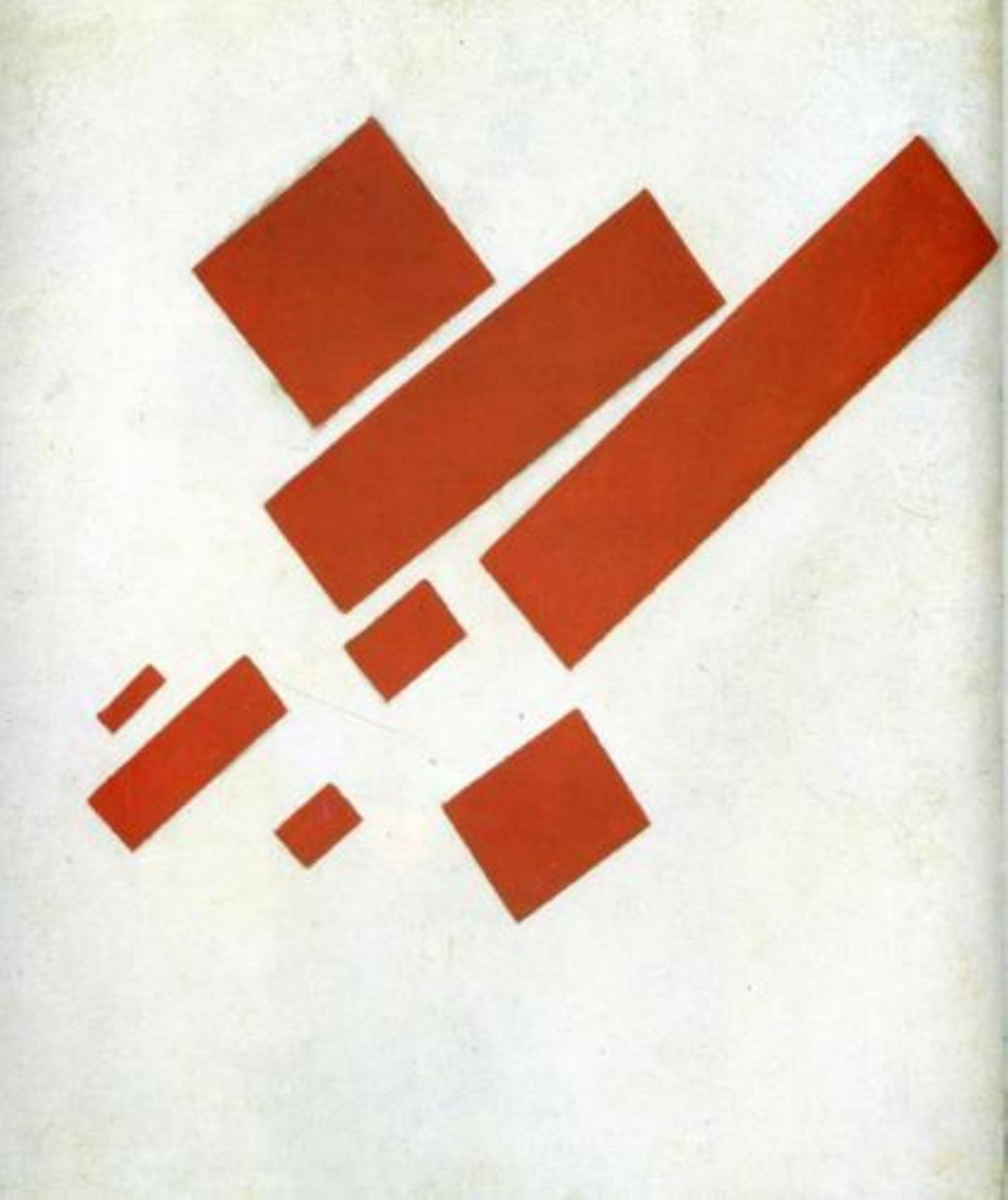} &\includegraphics[height=2.8cm, width=4cm]{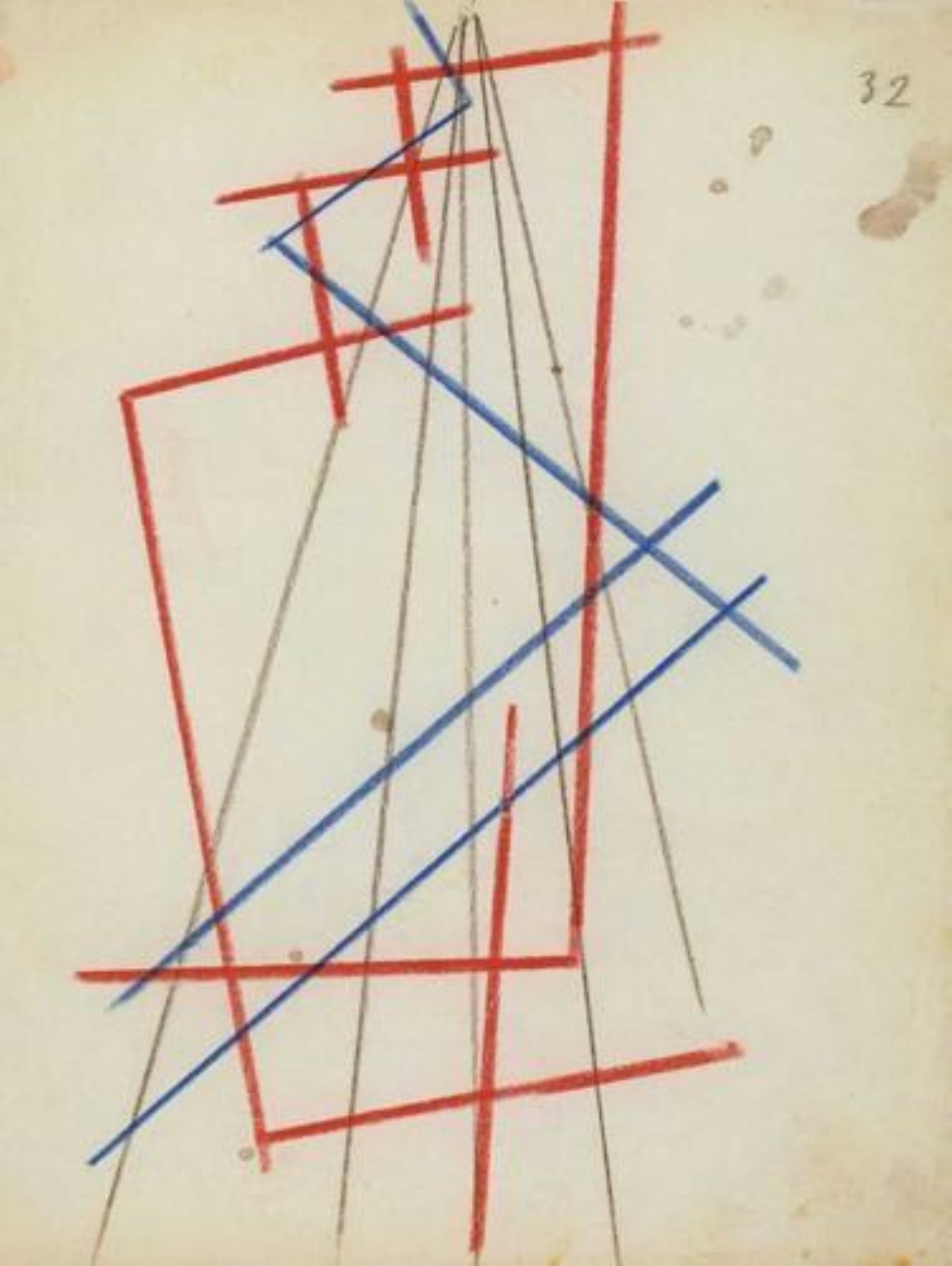} & \includegraphics[height=2.8cm, width=4cm]{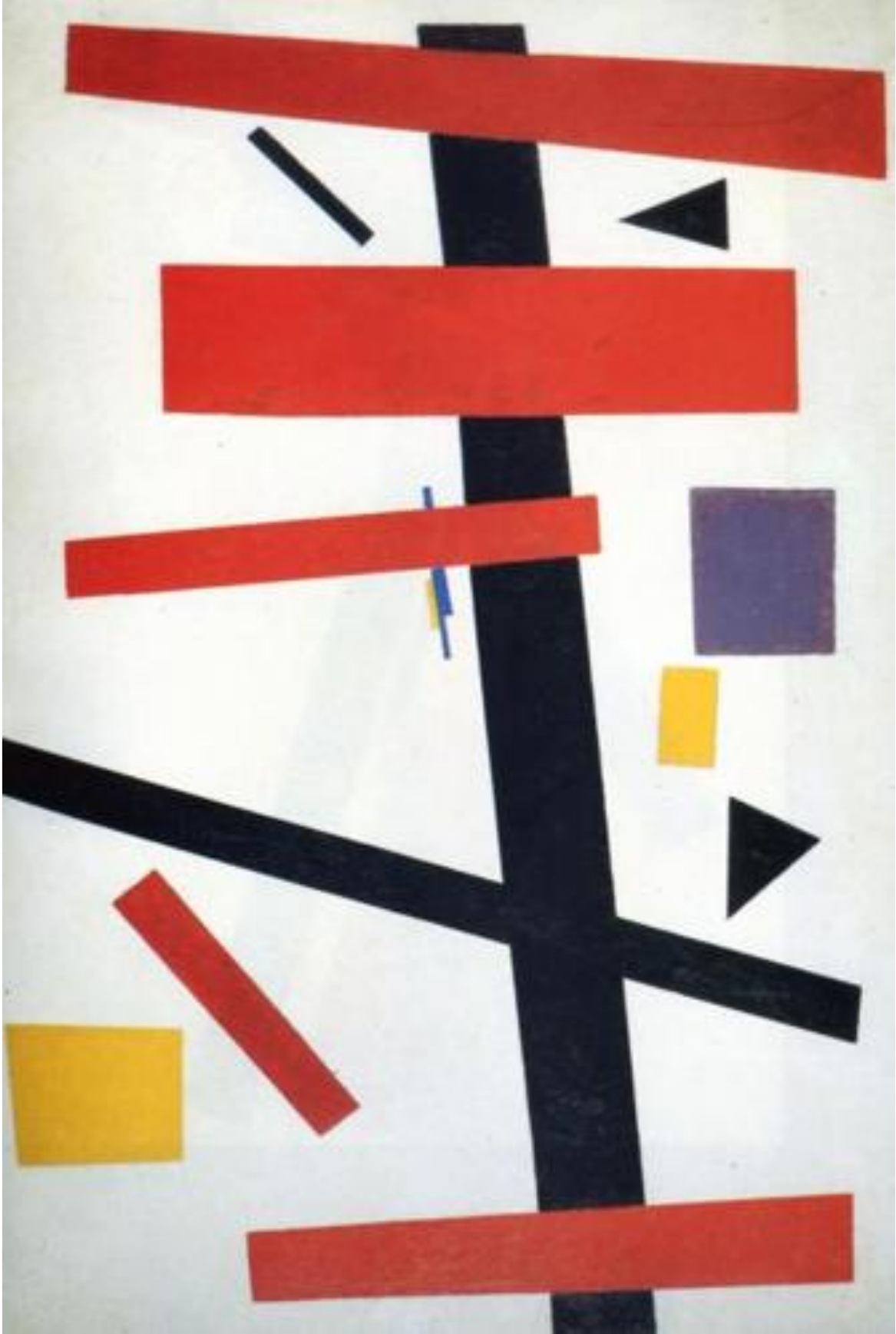}& \includegraphics[height=2.8cm, width=4cm]{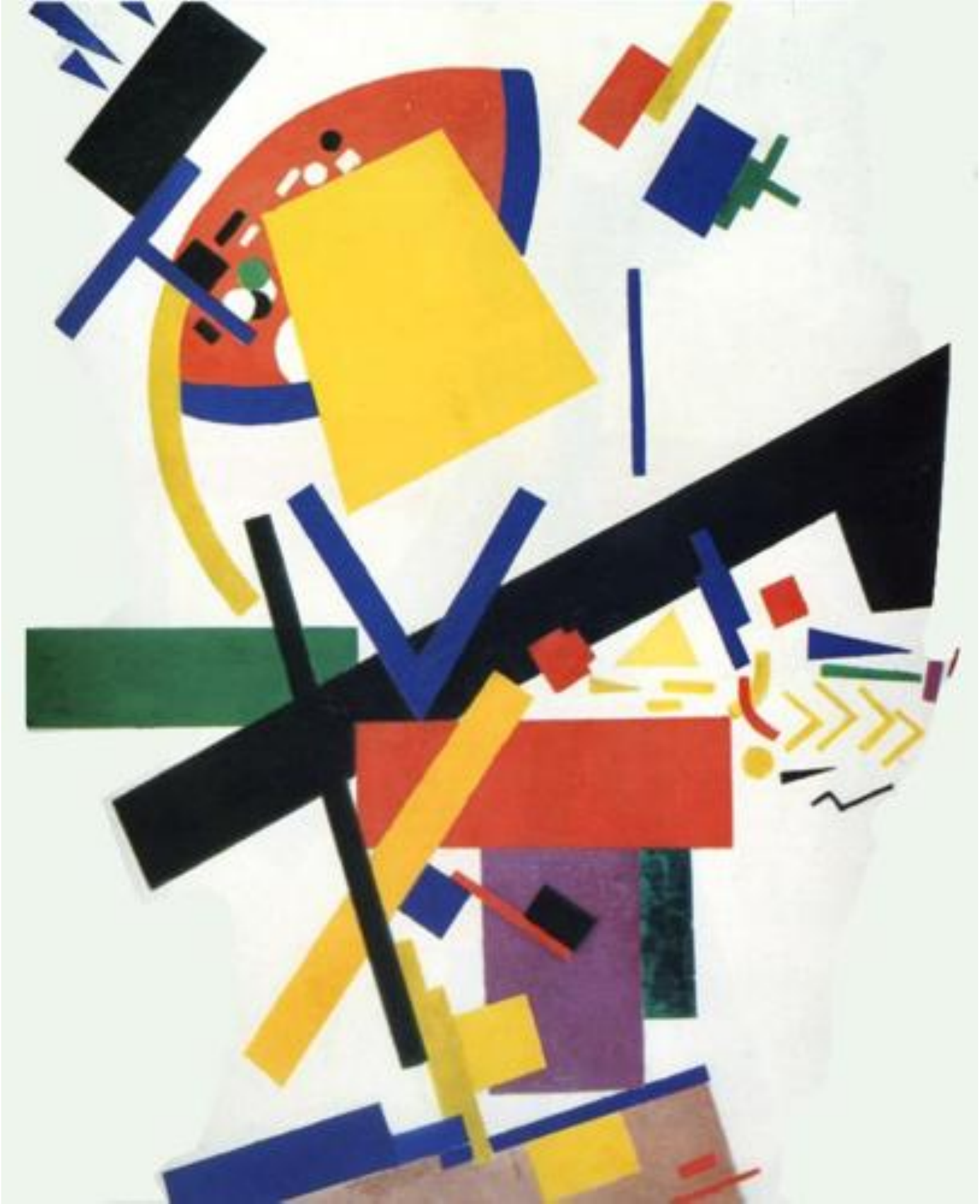} \\
\end{supertabular}
\end{center}

\end{table*}

Our dataset is the largest and most diverse open-source visual complexity dataset with 1,420 images in seven categories. In this section, we will first explain the image collection process and the different categories that we have used in our dataset. Next, we will discuss the data annotation using the Figure-Eight, a crowdsourcing platform.

\subsection{ Image Collection}
In our dataset, we have used images from seven diverse categories. Examples from each category in the increasing order of visual complexity are shown in Figure~\ref{image examples}.  Despite the connection between the problem of visual complexity and other problems in computer vision, there exists a gap between these topics. To overcome this gap, we have collected images from commonly-used datasets with the following categories:

{\bf Advertisement}: 200 images from the Advertisement dataset~\cite{hussain2017automatic}. Visual impression of advertisement plays a crucial role in economical competitions~\cite{pileliene3effect}. This category is selected in order to give ad designers insight into what factors impact the perceived complexity of an advertisement. Examples of factors that contribute to visual complexity of an ad are amount of text, size of text, importance of brand logo, and number and composition of various elements. The images in this category include both advertisements for products, and ads that campaign for/against something, e.g. for preserving the environment. 

{\bf Objects}: 200 images from the MSCOCO dataset~\cite{lin2014microsoft}. The purpose of this category is to understand how human perceives visual complexity of various objects and combination of objects. The number of objects is one of the leading factors contributing to the visual complexity of an image. This category can help researchers study the impact of characteristics of objects as well as the number of objects and their interaction with each other on visual complexity.

 {\bf Scene}: 200 images from the Places 2 dataset~\cite{zhou2017places}. The purpose of this category is to understand how humans perceive visual complexity of various scenes. It may facilitate the study of the roles of the image foreground and background in visual complexity analysis.

{\bf Interior Design}: We have collected 100 interior design images from the IKEA website~\cite{ikea}. This category is specifically selected to provide insight into how humans perceive the visual complexity of indoor spaces at home such as bedroom, living room, dining room, kitchen, and bathroom. Interior designers may want to understand how to design appealing interior spaces.

 {\bf Visualization and Infographics}: 200 images from the MASSViS dataset~\cite{borkin2013makes}. Visualization and infographics consists of charts, graphs, texts, and tables. Understanding the impact of each of these elements as well as their composition, thus, understanding the cognitive and perceptual processing of a visualization, can greatly influence the memorability, recognition, and comprehension of these designs.

{\bf Art}: 420 Artistic images from the PeopleArt dataset~\cite{westlake2016detecting}. This category consists of 10 images from each of the 42 categories of art styles and movements including Naturalism, Cubism, Socialist Realism, Impressionism, and Suprematism. Since the aesthetic beauty of an artistic image is directly influenced by the level of its visual complexity~\cite{eysenck1941empirical,reinecke2013predicting}, understanding the visual complexity of an artistic image can definitely help the artists to create more engaging artworks. 

{\bf Suprematism}: 100 images from the Suprematism category in the PeopleArt dataset for the analysis of geometric abstract art. The Suprematism category conveys various geometric shapes and objects in abstract form. This category enables studying the impact of various shapes, geometric objects, and composition on the perception of visual complexity.

\subsection{Dataset Groundtruthing}
\label{annotation}

\subsubsection{Pairwise Comparison Algorithm}

\vspace*{-0.2cm}

In order to obtain absolute ranking scores for an attribute of an image, in our case, visual complexity, one approach would be to ask users to assign a score to each image, where the score represents the ranking of the image relative to all other images. However, it has been shown that most people can only evaluate 5 to 9 options at a time.  
In addition, bias in the rating scale is a common problem in this type of groundtruthing~\cite{miller1956magical}.  

The use of pairwise comparison and conversion of the pairwise ranking to global ranking is a better alternative~\cite{arrow1950difficulty,david1963method,kendall1940method}.
Pairwise comparison is a relative measure that helps reduce bias from the rating scale. It is also invariant under monotone transformation of the rating values and depends only on the degree of relative difference between one option over the other in the pair~\cite{gleich2011rank}.
  
Note that, for example, for a set of $n=$200 images, including all of the pairs would result in  $\binom{n}{2}=19,900$ comparisons.
 
However, it is shown that for the pairwise comparisons, not all of the pairs are required in order to get the final global ranking, and information about a percentage of the pairs, $\ell \ll \binom{n}{2} $, is adequate~\cite{ChangYuWaAsFi16}. 

In many practical applications with partially observed measurements or budget constraints (e.g.\ PIB-PET scan for Alzheimer's disease detection), it is possible to use matrix completion methods in order to complement the results~\cite{candes2009exact,gleich2011rank,KimJaHwJoSi17}.

n this work, we follow the pairwise comparison approach.  Our algorithm is iterative and selects two images randomly from the set of images in a particular category in each step.  Images that have been selected in previous steps are less probable to be chosen in subsequent steps.  The algorithm terminates once a target number of comparisons is reached.   We decided on different target numbers for different categories, assuming that the visual complexity of images in some categories is easier to evaluate by human judgment than in others.  For the categories scenes, advertisement, visualization, and objects we decided to run our algorithm until $\ell=$4,000 pairs are found, which results in 40 comparisons per each image, on average, given that the these categories have $n=$200 images each.  For the interior design category, we ran the algorithm until $\ell=$2,000 pairs are found, also resulting in 40 comparisons per image, on average.  For the art category, we obtain $\ell=$14,700 pairs, which results in 70 comparisons, per image, on average.  Finally, for the Suprematism category, we used all possible $\ell =\binom{n}{2}=$ 4,950 pairs, which resulted in 99 comparisons per image.

\vspace*{-0.3cm}

\subsubsection{Crowdsourcing Methodology}

\begin{figure}[t!]
  \includegraphics[width=1\linewidth]{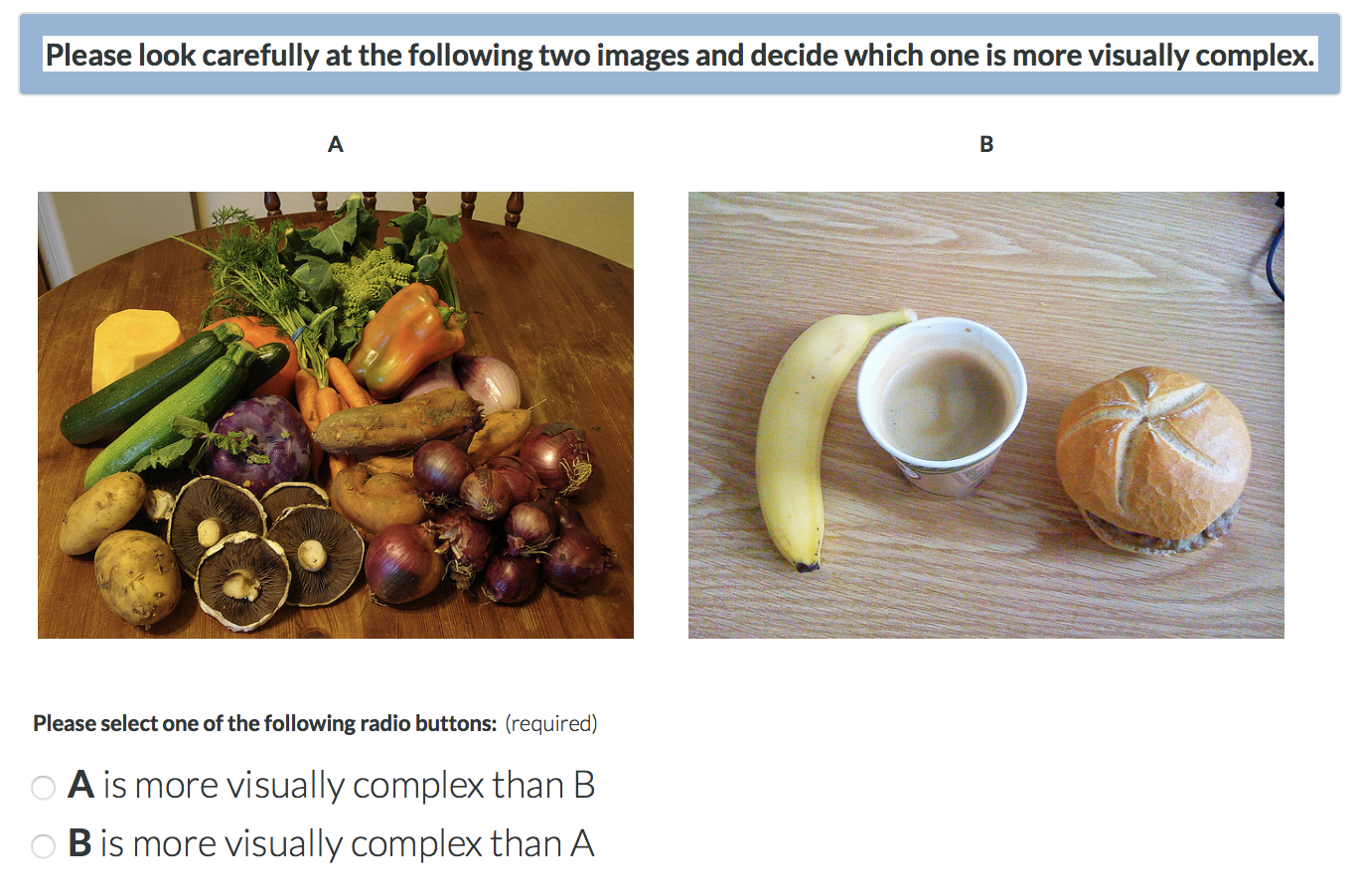}
  \centering
  \caption{\small Screenshot of one of the pairwise comparisons shown to Figure-Eight contributors for the objects category.}
  \label{fig:experiment}
\end{figure}

To minimize potential bias caused by raters, we collected our human judgments via crowdsourcing.
We used the Figure-Eight platform~\cite{FigureEight}.  Each task was distributed to five contributors. Contributors (also known as crowdworkers in Amazon Mechanical Turk) were shown ten pairs of images per page and asked which of the two images in each pair is visually more complex. We explained visual complexity by attributes such as cluttered background, numerosity and variety of objects, people, textures, patterns, and shapes.  We are using a forced-choice methodology, in which the contributors are supposed to select either image A or image B (Figure~\ref{fig:experiment}). In the case of similarly visually complex images, contributors are requested to select intuitively which image is more visually complex. 

Our contributors are selected from Figure-Eight ``level-3 contributors'' in Figure-Eight, who have shown to produce accurate answers in previous work.
Each contributor is shown 10 pairwise comparison task in a page and for each page they are paid \$0.10 regardless of their choices.  We did not allow any contributor to perform more than 300 tasks but did not select a lower bound for the number of tasks.
Workers were not restricted by geographical locations.

Test questions, geared towards quality control, were distributed to contributors randomly throughout the entire job to make sure they are attentive to the task. While all the comparison tasks were paid, only the answers from the contributors who maintain a passing score of 90\% or above on test questions were kept.  
\vspace*{-0.3cm}
\subsubsection{ Pairwise Ranking versus Global Ranking} \label{prvsgr}

\begin{figure*}[h]
\centering
\begin{subfigure}{0.22\textwidth}
\includegraphics[width=\linewidth]{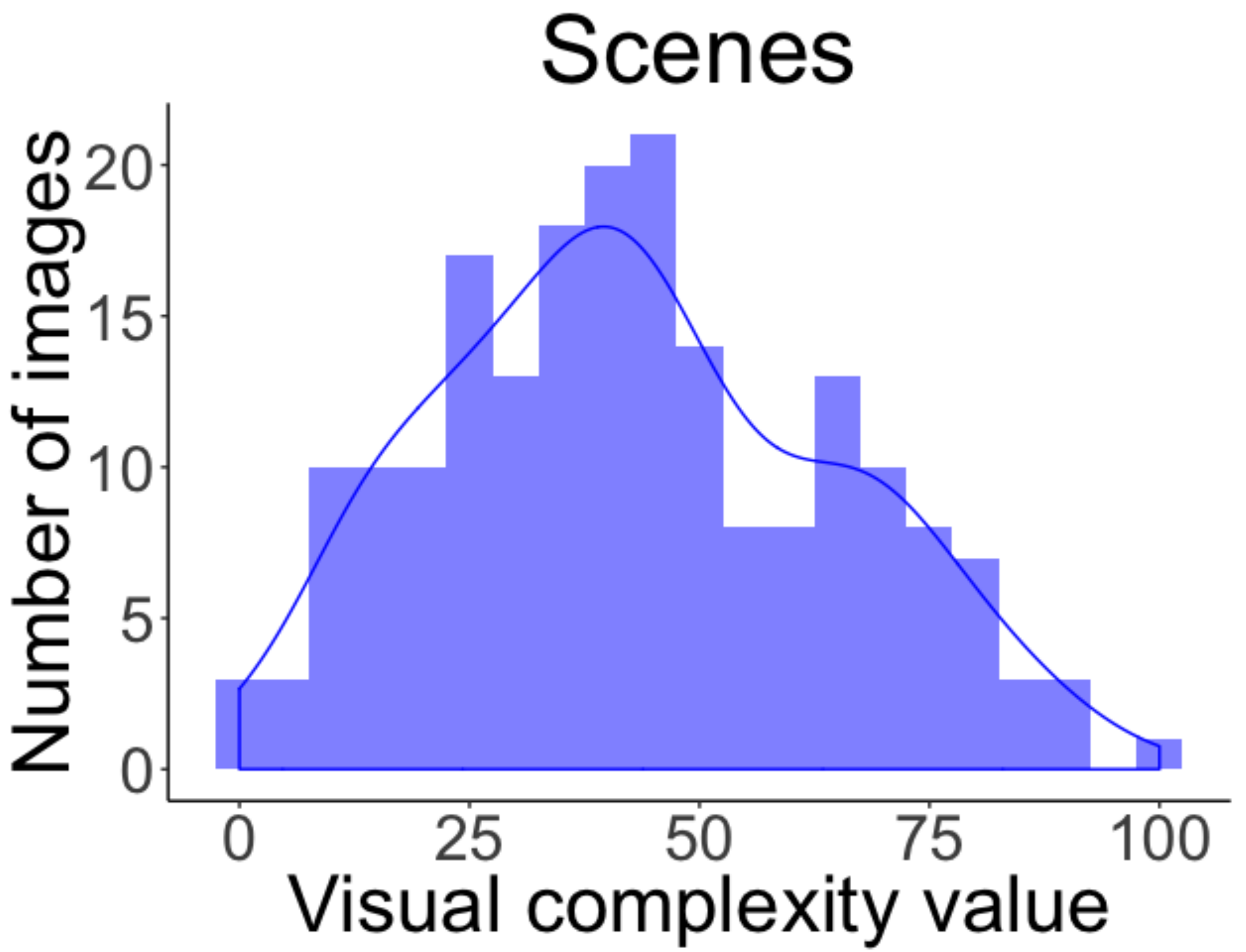}
\end{subfigure}
\begin{subfigure}{0.22\textwidth}
\includegraphics[width=\linewidth]{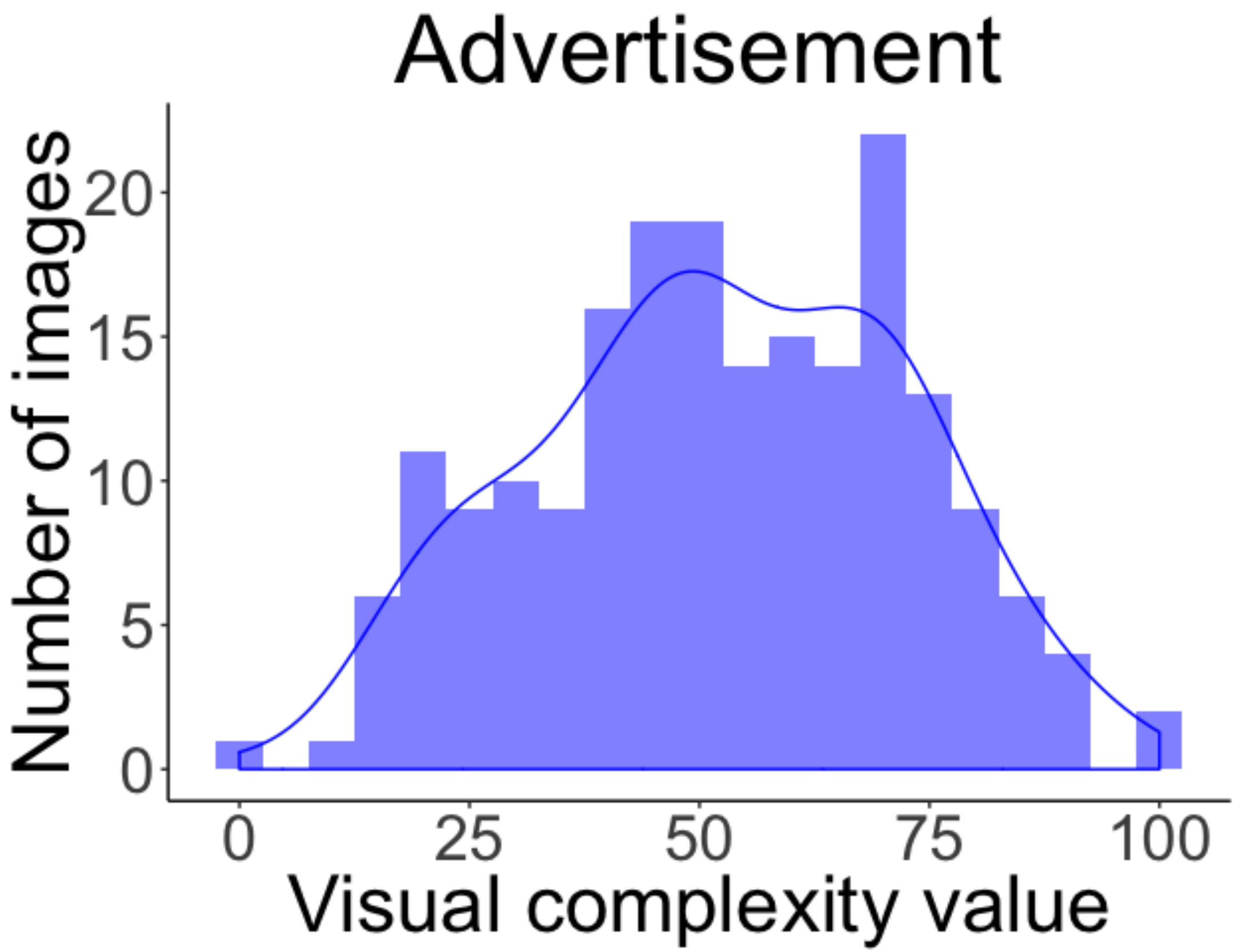}
\end{subfigure}
\begin{subfigure}{0.22\textwidth}
\includegraphics[width=\linewidth]{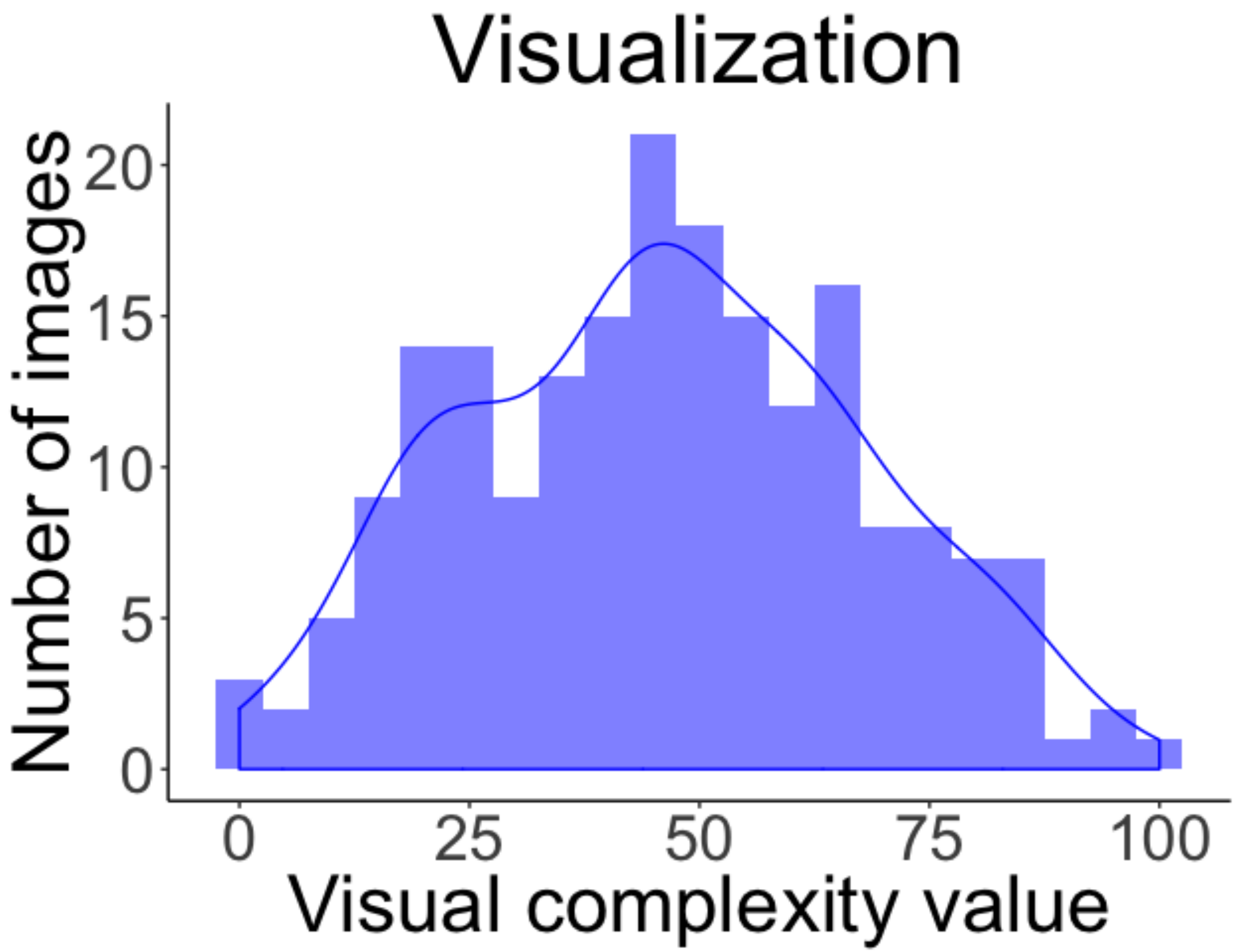}
\end{subfigure}
\begin{subfigure}{0.22\textwidth}
\includegraphics[width=\linewidth]{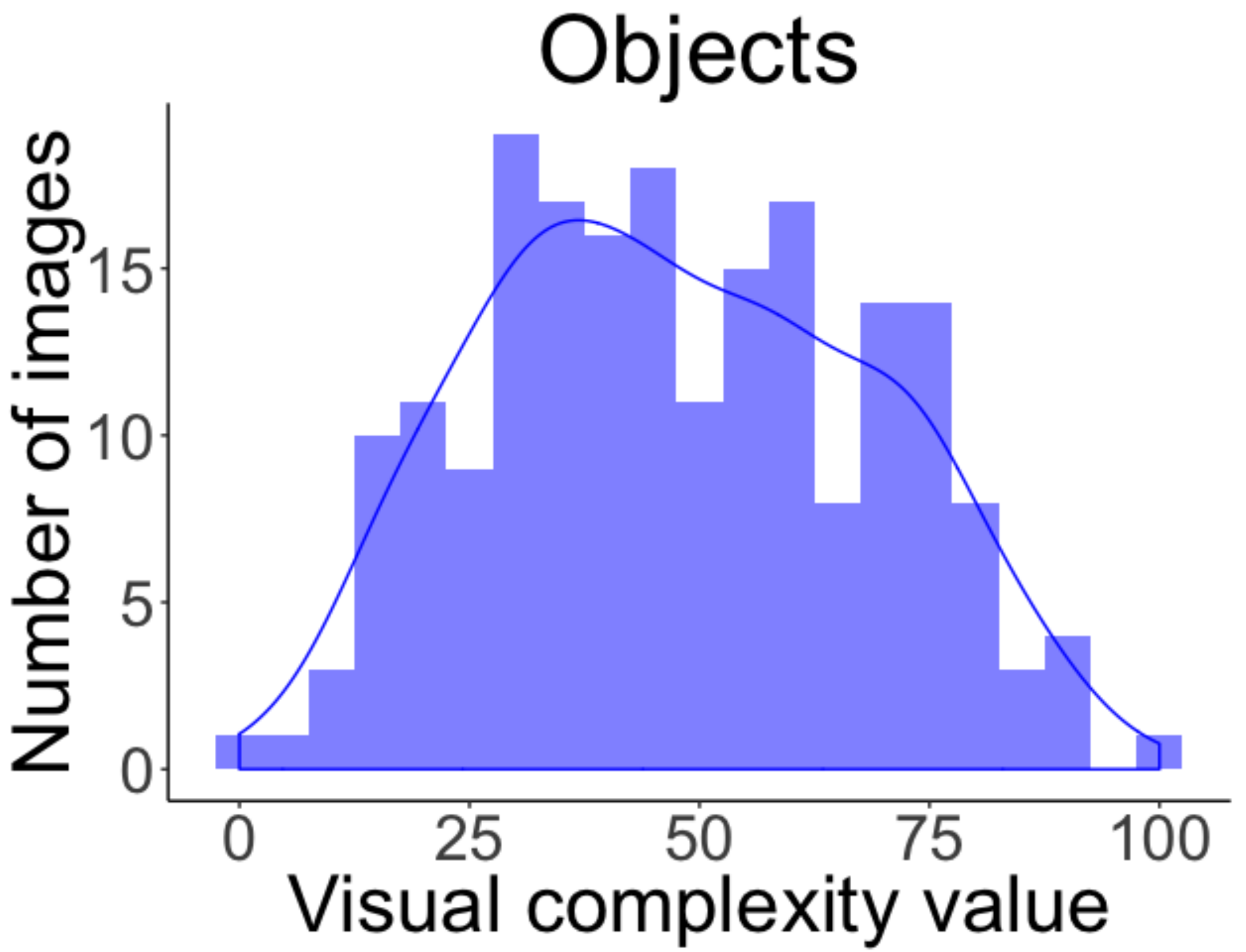}
\end{subfigure}

\medskip
\centering
\begin{subfigure}{0.22\textwidth}
\includegraphics[width=\linewidth]{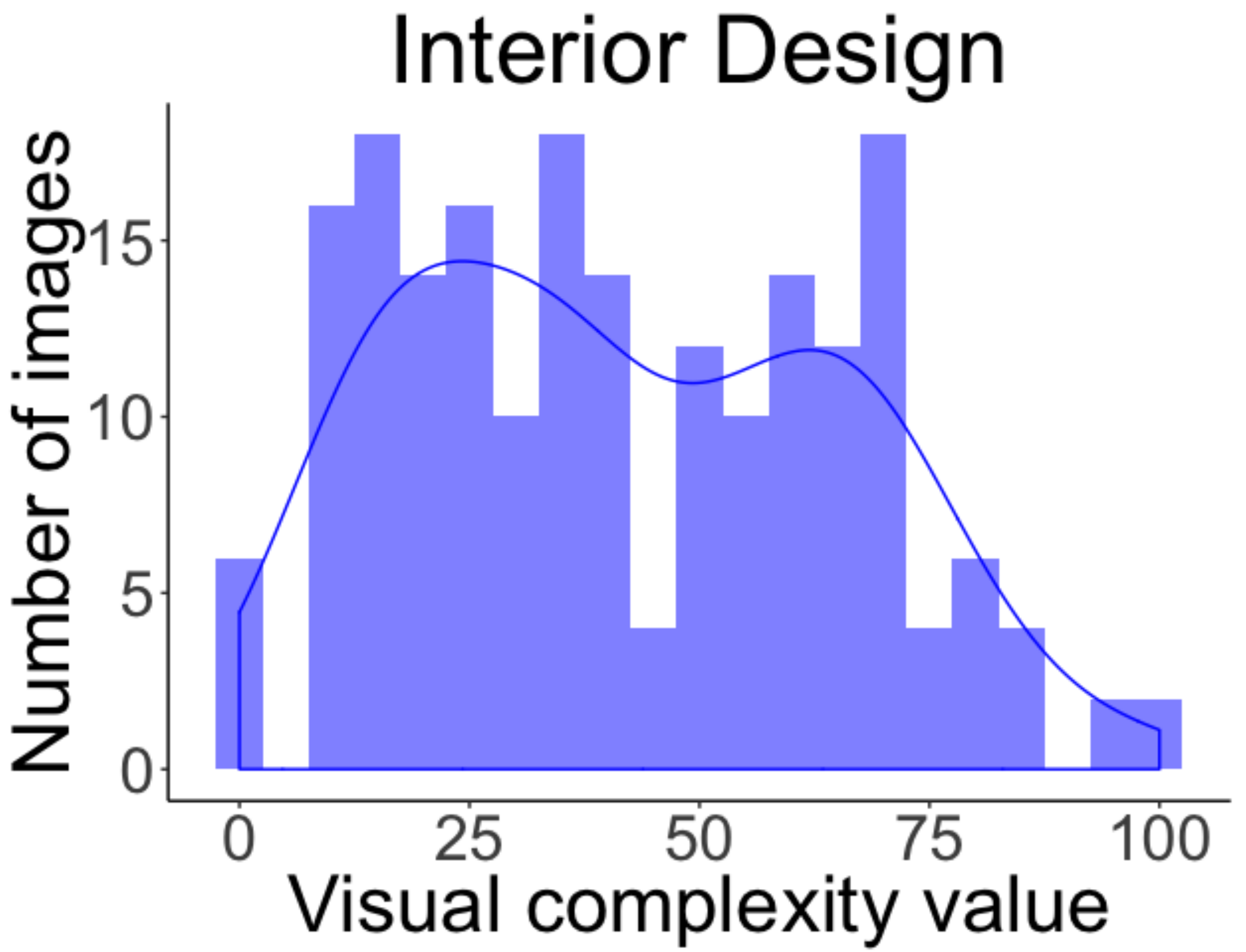}
\end{subfigure}
\begin{subfigure}{0.22\textwidth}
\includegraphics[width=\linewidth]{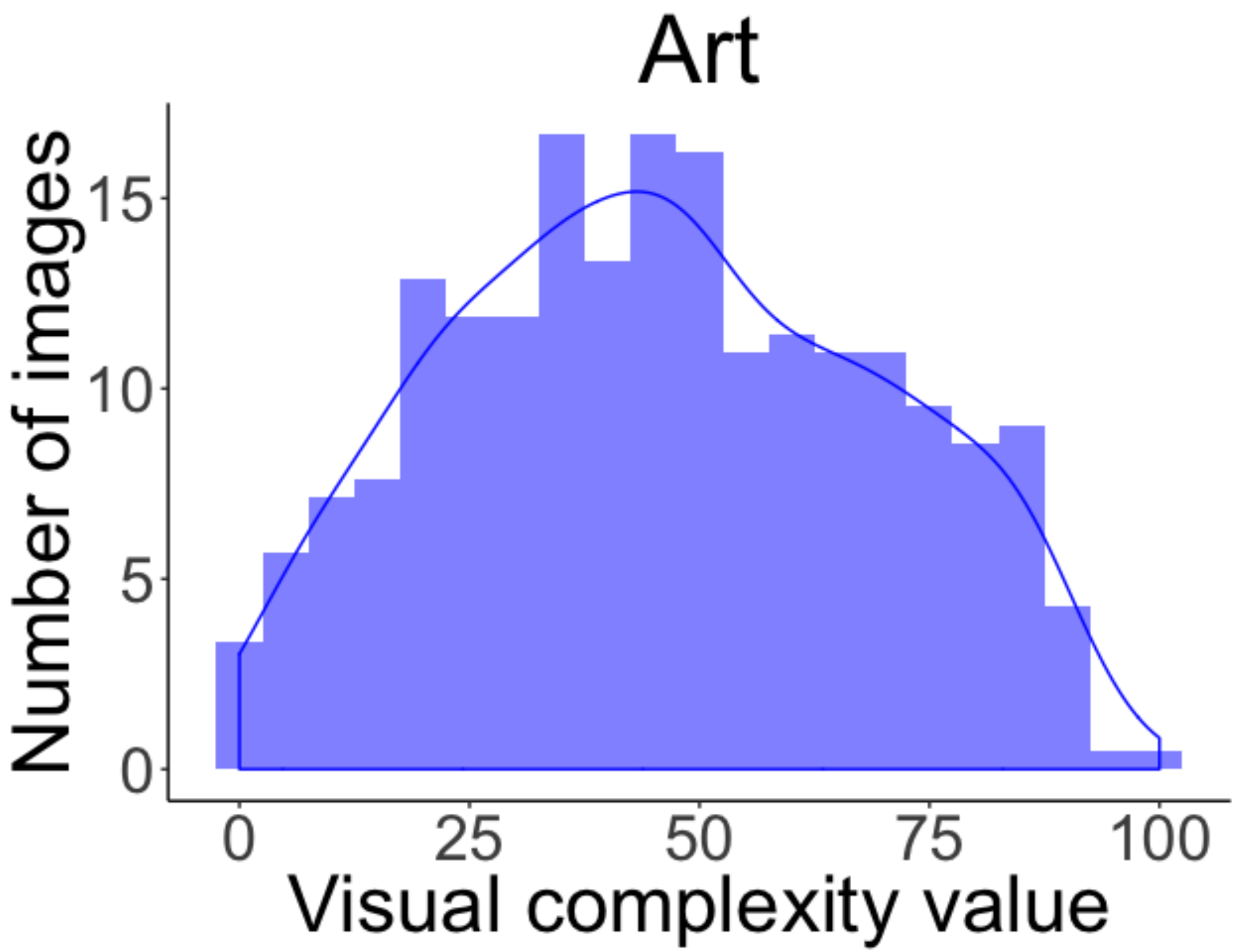}
\end{subfigure}
\begin{subfigure}{0.22\textwidth}
\includegraphics[width=\linewidth]{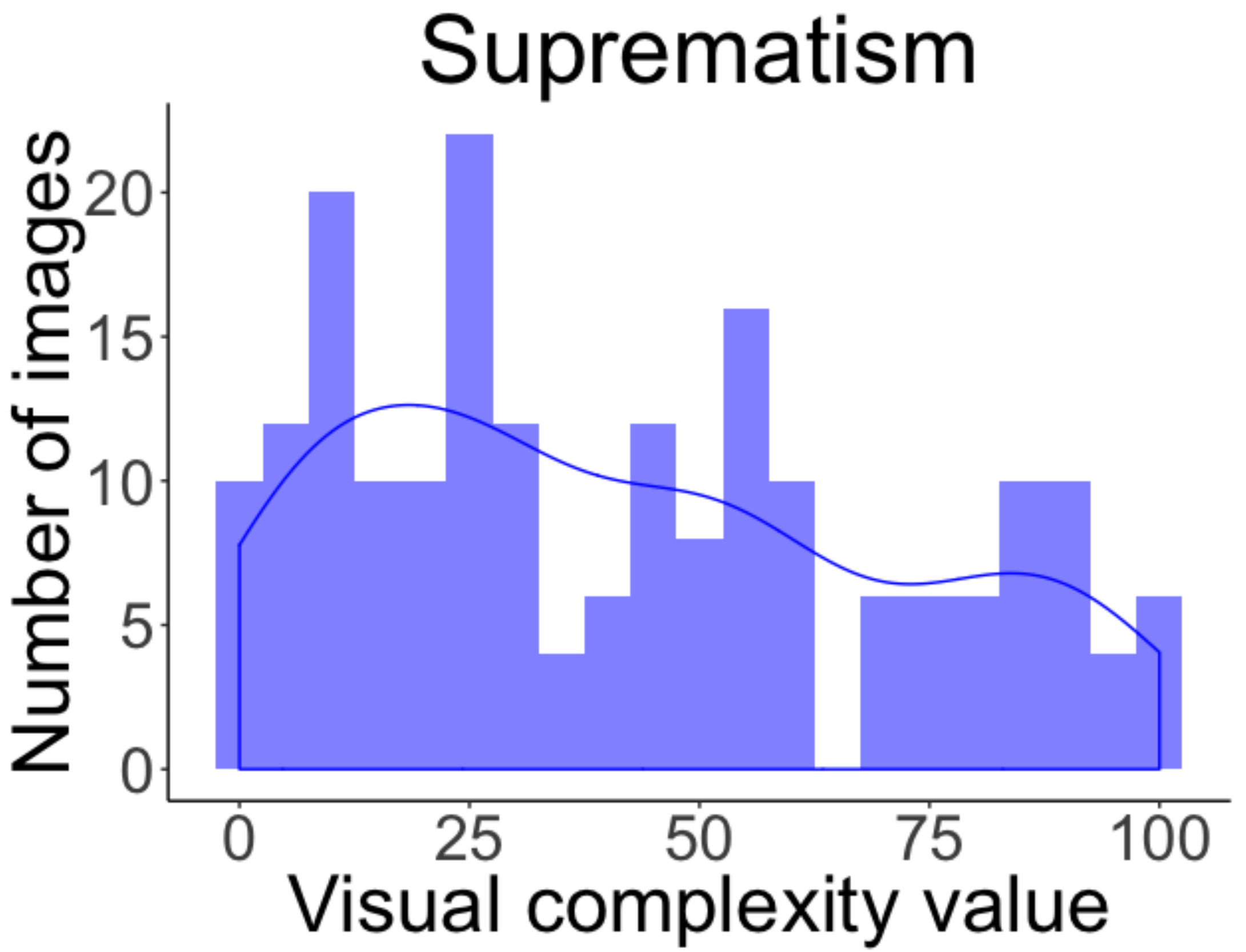}
\end{subfigure}

\caption{Distribution of absolute visual complexity scores per category for the {\sc Savoias} dataset.}
\label{histogtams}

\end{figure*}

In this work, we follow the pairwise comparison approach where the users on the crowdsourcing platform are asked to select whether image A or B is more visually complex. After the pairwise scores are collected, they need to be converted to absolute visual complexity scores. In order to convert pairwise ranking of images to global ranking, we applied two separate approaches, namely the Bradley-Terry method~\cite{bradley1952rank} and matrix completion. We then compare our results obtained from these two methods to validate our absolute scores.

We denote the pairwise comparison matrix as a count matrix $S = \{s_{i,j}\}$, where $s_{i,j}$ is the ratio of number of times that the contributors have selected image $i$ as more visually complex compared to image $j$ over the total number of times that image $i$ and $j$ have been compared. Thus, $s_{i,j} + s_{j,i} = 1$.
The problem here is to find $c_i$, the absolute score of image $i$.

The {\bf Bradley-Terry method}~\cite{bradley1952rank} describes the probability of choosing image $I_i$ over image $I_j$ as a Sigmoid function of the score difference between the two images, 
\begin{equation}
    P(I_i > I_j ) = F(\Delta_{i,j}) = \frac {e^{\Delta_{i,j}}}{1+e^{\Delta_{i,j}}},
\end{equation}
where $\Delta_{i,j} = c_i - c_j$. The score parameter $c$ can be estimated by solving a maximum a posteriori (MAP) problem, i.e., maximizing
\begin{equation}
    \log Pr(S|c) = \Sigma_{i,j} s_{i,j} F(\Delta_{i,j}),
\end{equation}
where the prior is a uniform distribution. This optimization problem can be solved using gradient descent~\cite{ChangYuWaAsFi16}.

{\bf Matrix Completion method} assumes, if $s_{i,j}$ is greater than $0.5$ (image $i$ is more visually complex than image $j$), $s_{j,k}$ is greater than $0.5$, and the pairwise comparison between image $i$ and $k$ is missing, using image $j$ as a link, we can infer that image $s_{i,k}$ is also greater than $0.5$. Now we can create matrix $\hat{S}$ by filling the missing elements of matrix $S$:
\begin{equation}
     \hat{s}_{i,k}=
    \begin{cases}
      s_{i,k} & \text{if}\ s_{i,k} \in S \\
      \frac{1}{m} \Sigma_{j=1}^m \frac{s_{i,j} + s_{j,k}}{2} & \text{else if}\ s_{i,j}\in S,~ s_{j,k} \in S~ \\ & ~\text{and} ~s_{i,j} > 0.5, s_{j,k} >0.5,
    \end{cases}
\end{equation}
where $m$ is the number of existing pairs of $s_{i,j}$ and $s_{j,k}$ in the count matrix $S$. 
For matrix completion, note the following points:
\begin{itemize}
\vspace*{-0.2cm}
    \item As indicated in the formula, we only consider $s_{i,j}~ \text{and} ~s_{j,k} \in S$ if they are greater than $0.5$. Therefore, if $s_{i,j} > 0.5$ and $s_{j,k} < 0.5$, we will not make any judgments about the missing pair $s_{ik}$.
    \vspace*{-0.2cm}
    \item For those pairs for which we have the result in one direction, we can fill the matrix in the other direction by this formula: $\hat{s}_{k,i} = 1 - \hat{s}_{i,k}$. 
    \vspace*{-0.2cm}
  
    \item For the rare case that a pair is not connected in either of directions, we use $\hat{s}_{i,k} = \hat{s}_{k,i} = 0.5$. 
    \vspace*{-0.2cm}

\end{itemize}

When the count matrix $\hat{S}$ is completed, the absolute score for each image is the mean of the pairwise scores for that image:
\begin{equation}
    c_i = \frac{1}{n} \sum_{j=1,~\hat{s} \in \hat{S}}^n ~\hat{s}_{i,j}.
\end{equation}

To confirm the correctness of the two aforementioned methods and our final scores, we evaluated the correlation between the global ranking scores obtained from these two methods. For all image categories, we obtained correlations higher than $0.98$ between the two methods.
We will report the results based on the Bradley-Terry method in Section~\ref{results-methodology}.

\begin{table*}[h]
\begin{center}
\small
\vspace*{-0.4cm}

    \begin{tabular*}{\textwidth}{l||@{\extracolsep{\fill}}ccccc}
         Model & \shortstack{Edge Density\\  ~\cite{rosenholtz2007measuring}} & \shortstack{Compression Ratio\\ ~\cite{corchs2014no}} & \shortstack{Number of Regions \\ ~\cite{comaniciu2002mean}} 
         & \shortstack{Feature Congestion\\ ~\cite{rosenholtz2007measuring}} & \shortstack{Subband Entropy\\ ~\cite{rosenholtz2007measuring}} \\
         \hline
         \hline
         Scenes & 0.16 & 0.30 & \textbf{0.57} & 0.42 & 0.16 \\
         Advertisement & 0.54 & \textbf{0.56} & 0.41 & \textbf{0.56} & 0.54 \\
         Visualization & 0.57 & 0.55 & 0.38 & 0.52 & \textbf{0.61} \\
         Objects & 0.28 & 0.16 & 0.29 & \textbf{0.30} & 0.10 \\
         Interior Design & 0.61 & \textbf{0.68} & 0.67 & 0.58 & 0.31 \\
         Art & 0.48 & 0.51 & \textbf{0.65} & 0.22 & 0.33 \\
         Suprematism & 0.18 & 0.60 & \textbf{0.84} & 0.48 & 0.39 \\

    \end{tabular*}

    \caption{
 \small Baseline results for the { \sc Savoias} dataset. We use the Pearson correlation coefficient to measure the performance of five algorithms. The object category is the most challenging category in our dataset -- the average correlation between ground-truth score and the score of the best performing algorithm (Feature Congestion) is only $0.30$.}
    \label{table:baseline}
    \end{center}
    
\end{table*}
\section{Results}
\label{results-methodology}
\setlength{\tabcolsep}{6.5pt}

In this section, we present an analysis of the visual complexity scores of {\sc Savoias}, discuss the results of the crowdsourced groundtruthing procedure, and describe the performance of five state-of-the-art algorithms on {\sc Savoias}.

\subsection{Distribution of the ground-truth scores}

Initial analysis of the distribution of the absolute scores showed that the absolute scores are mostly distributed around zero.
To mitigate this issue, we rescaled the range of pairwise scores, so that they are in the interval of $[0.33, 0.66]$ instead of $[0,1]$, while still maintaining $0.5$ as the score that represents equal visual complexity of an image pair.

Visual inspection of Figure~\ref{histogtams}, which presents the distribution of scores for the seven categories, shows that the rescaling step was successful -- each histogram is well distributed among the range of visual complexity numbers.

\begin{table}[t]
\small
\centering
\begin{tabular}{l||@{\extracolsep{\fill}}c@{\hskip 0.07in}c@{\hskip 0.07in}c@{\hskip 0.07in}c@{\hskip 0.07in}c@{\hskip 0.07in}c@{\hskip 0.07in}c@{\hskip 0.07in}c}
    \hline
   & 
   \begin{turn}{60}{Scenes}\end{turn} &
   \begin{turn}{60}{Ad.}\end{turn} & 
   \begin{turn}{60}{Vis.}\end{turn} &
   \begin{turn}{60}{Objects}\end{turn} & 
   \begin{turn}{60}{Interior}\end{turn} &
   \begin{turn}{60}{Art}\end{turn} &  
   \begin{turn}{60}{Sup.}\end{turn} &
   \begin{turn}{60}{Avg.}\end{turn} 
        \\
 \hline
 \hline
 Overall 
 \\ Satisfaction & 4.2 & 4.3  & 4.1& 4.5 &4.1 & 4.3 & \textbf{3.8} & 4.2 \\
 \hline
 Ease 
 \\ of Job & 4.1 & 4 & 4 & 4.3  & 3.9& 4.1 & \textbf{3.6} & 4 \\
\hline
\end{tabular}
\vspace*{-0.2cm}

\caption{\small Contributor satisfaction scores and ease of job scores for each of the crowdsourced datasets. Scores are reported out of 5 and provided by 237 internet workers. \vspace*{-0.3cm}
}
\label{Tab:satisfaction}
\end{table}

\subsection{Results on the Groundtruthing Methodology}

The number of crowdplatform contributors who provided judgments on visual complexity for {\sc Savoias} was 1,687.

Their overall satisfaction score was 4.2 out of 5 (Table~\ref{Tab:satisfaction}). The satisfaction score is a combination of ``instructions clear,'' ``test questions fair,'' ``ease of job,'' and ``pay.'' 

{\bf Validity of partial matrix versus full matrix comparison:}
 Here, we evaluate the accuracy of the absolute visual complexity scores as a function of $\ell$, number of pairwise comparisons between images. Since we have the full matrix comparison for the Suprematism category, we can perform such an analysis. 
Recall the notations from the Section~\ref{prvsgr}, where $S = \{s_{i,j}\}$ is the count matrix for the pairwise comparisons and $C = \{c_i\}$ is the list of absolute visual complexity scores, i.e., the output of the Bradley-Terry algorithm.
We define $S_\ell$ and $C_\ell$ as the count matrix and absolute scores where $\ell$ number of pairs have been selected for crowdsourcing. 

We define $S_f$ and $C_f$ as the full count matrix and resulting absolute scores, where $\ell = \binom{n}{2}$.

\begin{figure}[t]
\centering
 \includegraphics[ width=0.9\linewidth]{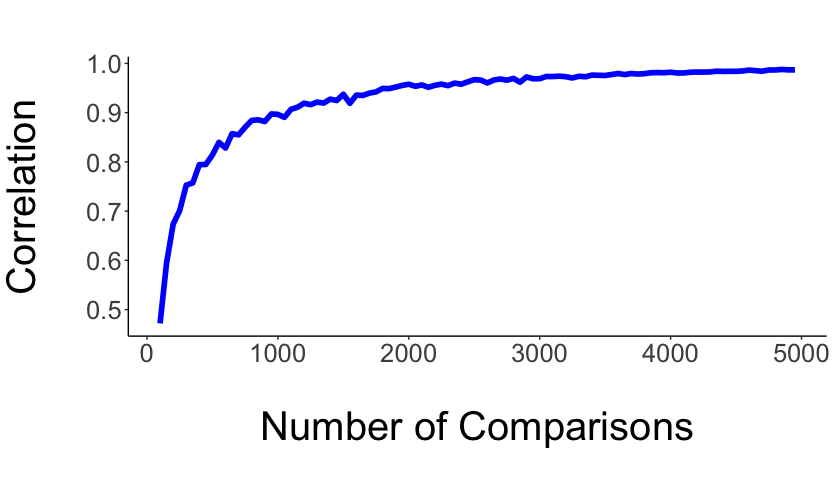}
\vspace*{-0.3cm}
\centering
  \caption{\small For the category Suprematism, the correlation between $C_\ell$ and $C_f$ is shown as a function of $\ell$.  High correlations can be achieved for some values of $ \ell \ll \binom{n}{2}$.
  }
  \label{corr_vs_k}
\vspace*{-0.3cm}

\end{figure}

The correlation between the visual complexity scores based on $C_\ell$ and $C_f$ for $\ell$ in the range of $[100-4950]$ is shown in Figure~\ref{corr_vs_k}, which highlights the trade-off between the accuracy and the number of pairwise comparisons.  
For example, if only 2,000 pairs had been chosen to define $S_\ell$ for the Suprematism category, the result would be close to the result of a full comparison, since the correlation between $C_l$ and $C_f$ is $0.96$.  Given this result for the Suprematism category, we hypothesize that high correlations can also be achieved for the other six categories if $\ell$ is selected to be much smaller than $\binom{n}{2}$.
 
\subsection{Baseline Results} 
We evaluated the performance of five state-of-the-art algorithms on our dataset and explored which algorithm is more suitable to be used for images in a certain category.
We used the Pearson correlation coefficient as the metric to evaluate the performance of each of these algorithms.
As shown in Table~\ref{table:baseline}, none of the algorithms can provide a correlation higher than $0.6$ for the objects, scenes, and advertisement categories. This highlights the need for a more sophisticated algorithm for categories of images that contain human-identifiable regions such as objects or people.

Furthermore, the best-performing algorithm is different for each category of images. This can be attributed to the diverse characteristics of each category.  A single algorithm that works well on all categories does not exist yet.
\subsection{Discussion}

Comparing the correlations between visual complexity scores based on crowdplatform contributors and the state-of-the-art algorithms, we observe that Suprematism, which is the most challenging category for the contributors, had the highest correlation for one of the algorithms. On the other hand, contributors found the object category to be the least challenging category, while all the algorithms performed poorly on this category (the highest correlation is only $0.3$).    Based on this observation, we postulate that the tested algorithms are more capable of making decisions based on the low-level features such as geometric shapes, textures, and patterns, found in the Suprematism category, than image features such as objects and people, which are easier for human contributors.

Furthermore, by analyzing the distribution of the ground-truth scores for each category visually (Figure.~\ref{histogtams}) and computationally (by estimating kurtosis), we observe that most of the categories have a distribution that resembles a Gaussian.  The Suprematism category, however, has a more uniform distribution (kurtosis = $1.915$).
This observation shows that there exist some images in the Suprematism category that are either very visually complex or not visually complex at all, and the contributors were able to distinguish them. For the other images in this category, it was not a trivial task for the contributors to recognize minor differences visual complexity, and so the they evaluated the ``ease of job'' to be low (Table~\ref{Tab:satisfaction}).

\section{Conclusions}
In this work, we introduced {\sc Savoias,} a new dataset for the analysis of visual complexity in images. {\sc Savoias} compromises of more than 1,400 images, which belong to seven diverse categories.  The ground-truth values were obtained by processing the judgments of 1,687 crowdplatform contributors who compared the visual complexity of more than 37,000 pairs of images. 

We suggest that the diverse range of image features found in {\sc Savoias} can support the development of new algorithms for quantifying visual complexity, and, moreover, be leveraged in research in other areas of computer vision, such as segmentation, object recognition and detection, visual search, image captioning, and visual question answering.  The most promising route might be to design algorithms that try to evaluate human-identifiable image features.
Furthermore, our proposed dataset can facilitate the research in the fields of psychophysics and cognitive science to find the underlying factors in the stimulus that affect the perception of visual complexity in humans.  Lastly, artists, Web and graphic designers, interior designers, and advertisers may eventually benefit from our dataset once better algorithms are designed for the categories relevant to their fields.

{\bf Acknowledgments}
We would like to thank Yifu Hu and Yi Zheng for preparing the images for the interior design category of our dataset.
This paper is based on work that was supported in part by the National Science Foundation (grants 1421943 and 1838193).

{\small
\bibliographystyle{ieee}
\bibliography{egbib}
}

\end{document}